\documentclass[10pt]{article} 
\usepackage[preprint]{tmlr}

\usepackage{amsmath,amsfonts,bm}

\def\figref#1{figure~\ref{#1}}
\def\Figref#1{Figure~\ref{#1}}
\def\twofigref#1#2{figures \ref{#1} and \ref{#2}}

\def\eqref#1{equation~\ref{#1}}

\def\1{\bm{1}}

\DeclareMathAlphabet{\mathsfit}{\encodingdefault}{\sfdefault}{m}{sl}
\SetMathAlphabet{\mathsfit}{bold}{\encodingdefault}{\sfdefault}{bx}{n}

\usepackage{hyperref}
\usepackage{url}
\usepackage[acronym]{glossaries}
\usepackage{subcaption}
\usepackage{graphicx}
\usepackage{xcolor}
\usepackage{placeins}
\usepackage{booktabs}

\newacronym{sac}{SAC}{soft actor-critic}
\newacronym{kl}{KL}{Kullback-Leibler}
\newacronym{dsr}{DSR}{dynamical system reconstruction}
\newacronym{ds}{DS}{dynamical system}
\newacronym{rnn}{RNN}{recurrent neural network}
\newacronym{rl}{RL}{reinforcement learning}
\newacronym{alrnn}{ALRNN}{Almost Linear Recurrent Neural Network}
\newacronym{lsr}{LSR}{linear subregion}
\newacronym{pca}{PCA}{Principal Component Analysis}
\newacronym{mbrl}{MBRL}{model-based reinforcement learning}
\newacronym{lce}{LCE}{Learning Control Embeddings}
\newacronym{fp}{FP}{fixed point}
\newacronym{pwl}{PWL}{piecewise linear}
\newacronym{mdp}{MDP}{Markov decision process}
\newacronym{ffn}{FFN}{feedforward network}
\newacronym{mpc}{MPC}{model predictive control}
\newacronym{mrs}{MRS}{middle, ring and small}
\newacronym{svd}{SVD}{singular value decomposition}

\title{Learning Transferable Policies from Action-free Time Series Through Dynamical Embeddings}

\author{\name Niklas Emonds \email niklas.emonds@iwr.uni-heidelberg.de \\
    \addr 
    Hector Institute for AI in Psychiatry (HITKIP) \& Department of Psychiatry and Psychotherapy, Central Institute of Mental Health (CIMH), Medical Faculty Mannheim, Heidelberg University\\
    Interdisciplinary Center for Scientific Computing (IWR), Faculty of Mathematics and Computer Science, Heidelberg University
      \AND
      \name Georgia Koppe \email georgia.koppe@iwr.uni-heidelberg.de \\
      \addr IWR, Faculty of Mathematics and Computer Science, Heidelberg University\\
       HITKIP \& Department of Psychiatry and Psychotherapy, CIMH, Medical Faculty Mannheim, Heidelberg University\\
       Hertie Institute for AI in Brain Health, University of Tübingen, Germany
}

\def\month{MM} 
\def\year{YYYY} 
\def\openreview{\url{https://openreview.net/forum?id=XXXX}}

\begin{document}

\maketitle

\begin{abstract}
 Learning control from action-free recordings is challenging because intervention effects are unobserved and policies may exploit errors in reconstructed dynamics.
 We present a hierarchical model-based reinforcement learning framework that uses shared structure across related systems to learn system-specific control policies from action-free recordings. A hierarchical dynamical system reconstruction model captures shared dynamics and individual variation through low-dimensional embeddings. These embeddings are then reused to parameterize shared policy and value networks, linking differences in reconstructed dynamics to differences in control. Policies are trained entirely via simulation under an explicit intervention model with additive latent perturbations. Piecewise-linear recurrent neural networks enable mechanistic analyses of the controlled dynamics, while decoder-based constraints make the immediate effects of interventions interpretable in observation space and permit interventions on one modality while protecting another from direct manipulation. 
  On Lorenz-63 and double-pendulum systems, hierarchical policies improve transfer over independently trained policies. On Lorenz-63, they also achieve a higher mean reward than repeated planning with the same reconstructed models, perform comparably to methods trained with controlled interactions, and generalize to systems absent from policy training after embedding inference alone. Applications to neural–behavioral recordings demonstrate suppression of predicted movement under constrained neural perturbations. Together, these findings show how shared dynamical representations support transferable control and mechanistic hypothesis generation from action-free recordings.
\end{abstract}

\section{Introduction}
Controlling nonlinear \glspl{ds} is central to many fields of science and engineering. In neuroscience, control may involve selecting inputs that steer neural dynamics toward a desired state or behavioral outcome. Yet limited data and safety or ethical constraints often restrict opportunities to experimentally probe the effects of such inputs \citep{Krauss2021,Chandrabhatla2023-xm}. Neural and behavioral recordings may nevertheless be available across individuals or repeated sessions, providing information about shared dynamics and individual variation. We call these recordings \emph{action-free} when they contain no responses to the control actions considered here, even though task cues and behavior may be recorded. We study how to learn control policies from these recordings using reconstructed dynamics and an explicitly specified action model, \emph{without} further interaction with the underlying systems during training.

Having a model of the dynamics provides a basis for designing control policies. Two major approaches use such a model to select actions: repeated planning during deployment and prior learning of an explicit policy that maps the current state to an action. \Gls{mpc} follows the first approach, using the model to optimize a finite sequence of actions at each time step, apply its first action, and repeat the optimization from the next observed state \citep{RICHALET1978413,MAYNE2000789}. While \gls{mpc} relies on computationally heavy trajectory optimization at every step, \gls{mbrl} can amortize this cost by using the transition model to train a policy \citep{SUTTON1990216, moerland2022modelbasedreinforcementlearningsurvey}. In particular, latent world models allow a policy to be trained entirely on simulated trajectories and deployed with minimal latency \citep{pmlr-v97-hafner19a,hafner2020dreamcontrollearningbehaviors, Hafner2025}.
Although policy learning occurs in simulation, these methods typically learn action effects from controlled interactions with the real system. 

Learning from action-free recordings creates two difficulties. First, policies optimized in a learned simulator may exploit reconstruction errors, particularly when actions move the system into regions poorly constrained by the recordings. Without recorded control actions and their outcomes, there is no empirically supported intervention behavior toward which the policy can be regularized \citep{NEURIPS2020_a322852c, NEURIPS2021_f29a1797}. Reliable dynamics reconstruction is therefore essential for generating the long simulated trajectories on which policy learning depends, alongside regularization that limits exploitation of model errors. This is particularly challenging for chaotic systems, where small errors can amplify rapidly \citep{DeterministicNonperiodicFlow}. Second, even a model that perfectly reproduces the system’s dynamics without intervention does not tell us how a physical actuator would change its state. This relationship must therefore be specified through prior knowledge or an explicit modeling assumption.

To address the reconstruction problem, we build on hierarchical \gls{dsr}. \Gls{dsr} seeks models that recover attractor geometry and characteristic temporal structure beyond short-term prediction accuracy, and excels particularly at learning dynamical systems \citep{Durstewitz2023, pmlr-v202-hess23a}.
Hierarchical \gls{dsr} shares parameters across related systems while representing individual variation through low-dimensional embeddings, improving data efficiency and generalization when per-system data are limited \citep{ICLR2025_d4c96180,vermanimeta}.

We extend this hierarchy to control by reusing reconstruction embeddings to parameterize shared policy and value networks, trained with \gls{sac} entirely on simulated transitions \citep{pmlr-v80-haarnoja18b}. This incorporates learned dynamical structure into control, with differences between related systems determining corresponding variations within a shared policy family. In the absence of an empirical action distribution toward which policies could be regularized, hierarchical parameter sharing therefore provides an alternative constraint. It reduces the number of trainable parameters and limits specialization to individual world models.
Moreover, the controller can be applied to systems not encountered during policy training. Fitting a new embedding to their uncontrolled recordings yields a system-specific policy while shared actor and critic parameters remain fixed.

The second problem remains unresolved by observational data alone, since we cannot identify how a physical actuator couples to the system. Here, we therefore ask what changes in \emph{measured} variables could achieve a desired outcome. To study this question, we model interventions as additive actions that additively perturb the latent state, applied before the learned transition. Linear decoders make their immediate effects explicit in observation space, allowing us to characterize which variables the controller changes and by how much. They also permit modality-specific constraints. For example, interventions can alter decoded neural activity while preserving the immediate behavioral output, so that subsequent behavioral changes arise through the reconstructed dynamics \citep{https://doi.org/10.1002/hbm.70600}.

To understand how these changes achieve control, we use \glspl{alrnn} with piecewise-linear latent dynamics \citep{10.1371/journal.pcbi.1005542,10.1371/journal.pcbi.1007263,NEURIPS2024_40cf2729}. Analytic access to
fixed points and stability properties allows us to examine how interventions shift dynamical mechanisms.  
Together, these analyses generate mechanistic hypotheses about which changes in observed variables can achieve control and the dynamical mechanisms through which they act.

In sum, our contributions are:
(i) a formulation for learning control policies from action-free recordings that combines \gls{dsr} with \gls{mbrl} under an explicit latent intervention model;
(ii) a hierarchical controller that reuses reconstruction embeddings to share \gls{sac} across related \glspl{ds}, enabling adaptation to unseen systems without retraining;
(iii) decoder-based action constraints that make interventions interpretable in observation space and permit control across distinct intervention and target modalities;
(iv) a mechanistic analysis of learned control strategies using regional fixed points and local stability properties; and
(v) experiments demonstrating improved transfer to the underlying Lorenz-63 and double-pendulum systems relative to independently trained \gls{sac} controllers, embedding-based adaptation to unseen Lorenz-63 systems, and in-silico control of behavior in models fitted to neural–behavioral recordings.

\section{Related Work}
\textbf{Model-predictive control}
\Gls{mpc} is a standard framework for controlling nonlinear systems by repeatedly solving finite-horizon optimization problems under
dynamical and input constraints \citep{RICHALET1978413,MAYNE2000789}. For nonlinear and hybrid systems, piecewise-affine formulations provide an important
compromise between expressivity and tractability, since nonlinear dynamics can be represented through collections of locally affine regimes with region-dependent
control laws \citep{https://doi.org/10.1002/rnc.817}. This is conceptually close to the \gls{pwl} latent dynamics used in our work. However, classical \gls{mpc} typically assumes a reliable controlled model
and often relies on online optimization at deployment time. Recent learning-based extensions retain the receding-horizon principle while relaxing the requirement
of analytically specified dynamics: CEM-\gls{mpc} optimizes action sequences by sampling and refining them in a learned predictive model, whereas TD-\gls{mpc} jointly
learns a latent dynamics model and value function from controlled interaction data and performs short-horizon latent-space planning with terminal
value estimates \citep{pmlr-v155-pinneri21a, pmlr-v162-hansen22a}. Our CEM-\gls{mpc} baseline uses the same reconstructed \gls{alrnn} as the proposed method but plans anew at every time step. 

\textbf{Model-based and offline RL}
\Gls{mbrl} learns a world model and uses it for planning or policy optimization \citep{SUTTON1990216,moerland2022modelbasedreinforcementlearningsurvey}. Latent world models such as the Dreamer framework show that compact state-space
models can support policy learning through simulated rollouts \citep{pmlr-v97-hafner19a,hafner2020dreamcontrollearningbehaviors,Hafner2025, hafner2025trainingagentsinsidescalable}. These
methods deal well with high-dimensional observations, but usually assume either online interaction or sufficient logged state-action data. Offline
\gls{mbrl} addresses the resulting distribution-shift problem by penalizing model rollouts or policies that move away from the empirical state-action
distribution \citep{NEURIPS2020_a322852c,NEURIPS2021_f29a1797}. In our setting, however, no actions are observed, so there is no measured action
distribution against which policy behavior can be regularized. We therefore use hierarchical parameter sharing as an implicit regularizer.
Multitask and hierarchical \gls{rl} exploit shared structure across related control problems by sharing policy and value function parameters or
conditioning them on task embeddings, thereby improving data efficiency and generalization.
Hierarchy here refers to a parametric decomposition into shared and subject-specific components, not to temporal scales, levels of abstraction or option-based control \citep{NIPS1992_d14220ee, Barto2003,10.5555/3298483.3298491}.

\textbf{Reconstruction of families of dynamical systems}
\gls{dsr} seeks generative models that learn the true underlying \gls{ds} by reproducing attractor geometry and long-run temporal structure from time series \citep{Durstewitz2023, durstewitz2026position}. \glspl{alrnn} retain a \gls{pwl} structure with a parsimonious symbolic code, making local dynamical analysis feasible \citep{NEURIPS2024_40cf2729, brenner2026uncovering}. Hierarchical \gls{dsr} represents a family of systems through shared parameters and low-dimensional embeddings, which can improve estimation from limited per-system data and enable adaptation to unseen systems \citep{ICLR2025_d4c96180, vermanimeta}. We extend this framework from reconstruction to control by conditioning the policy and value function on the same embedding.

\textbf{Control in neuroscience and cross-modal interventions}
Adaptive neurostimulation provides a challenging application that motivates our setting. Common
methods optimize control policies directly, rather than learning a world model of the neural dynamics \citep{Chandrabhatla2023-xm,Krauss2021,Horn2025}. Closed-loop, personalized stimulation has to be designed under safety and data constraints that limit collecting interventional data \citep{ravivarapu2025sampleefficientreinforcementlearningcontroller}. Recent work therefore learns stimulation policies in simulated neurophysiological environments \citep{10.1145/3711896.3737397}, or fits state-transition models to observational recordings and applies control theory to them for mechanistic insight \citep{https://doi.org/10.1002/hbm.70600}. In each case the effect of an action is supplied by a simulator or by an assumed perturbation model, as it is here. Our neural experiments constrain that assumption further: actions are restricted to neural decoder directions, so any change in predicted behavior must pass through the learned latent dynamics.

\section{Methods}
\subsection{Preliminaries and problem setting}
\gls{mbrl} requires an environment model $p(z_{t+1},r\vert z_t,a_t)$. We define the reward $r$ by penalizing distance from a target state. Our goal is therefore to construct the transition dynamics $p(z_{t+1}\vert z_t, a_t)$ from short recordings of related systems in which no control actions $a_t$ were observed. We learn each system's dynamics without control actions, specify how actions perturb its state, and combine these components into an environment for policy learning.

\subsection{Learning the dynamics from recordings}
We first learn a model that can generate trajectories of each system without control, using established methods for \gls{dsr}. An \gls{alrnn} describes the latent dynamics, observation models connect the latent state to the recordings, and a hierarchical parameterization shares information across related systems \citep{NEURIPS2024_40cf2729,pmlr-v235-brenner24a,ICLR2025_d4c96180}. Together, these components provide the environments in which we train the policy (see also \figref{fig:Methods_1}).

\paragraph{State transitions} For subject $j$, an encoder maps the observation trajectory to latent states $z_t^{(j)}\in\mathbb{R}^L$. The \gls{alrnn} then advances this state according to
\begin{align}
z_{t+1}^{(j)} = F_{\varphi^{(j)}}(z_{t}^{(j)}, s_{t}^{(j)}) = A^{(j)}z_{t}^{(j)} + W^{(j)} \sigma(z_{t}^{(j)}) + C^{(j)} s_{t}^{(j)}  + h^{(j)},
\end{align}
where $A \in \mathbb{R}^{L\times L}$ is a diagonal self-connection matrix, $W \in \mathbb{R}^{L\times L}$ contains cross-connections, and $h \in \mathbb{R}^L$ is a bias term. The function $\sigma$ applies a ReLU to $P$ of the $L$ latent coordinates and leaves the others unchanged.
This yields a \gls{pwl} vector field that partitions latent state space into $2^P$ subregions within which dynamics are linear. The optional input $s_t\in\mathbb{R}^{K}$ and its weight $C\in\mathbb{R}^{L\times K}$ represent recorded external signals, such as task cues; it is distinct from the action introduced below. When no external inputs are modeled, this input term is omitted. 

Observation models decode the latent state into the observations $x_{t}^{(j)}\in\mathbb{R}^O$. Only a subset of these latent dimensions is decoded. When recordings contain several modalities, such as neural activity and behavior, each modality has its own decoder but shares the same latent dynamics. This lets us learn from the recorded variables and later define control objectives separately in those same variables.

\paragraph{Sharing information across systems} A short recording may not sufficiently constrain a model fitted to one subject alone. We therefore represent subject-specific parameters through shared group-parameters and subject vectors.
Specifically, subject-specific parameters $\varphi^{(j)}\in\mathbb{R}^{\vert \varphi^{(j)}\vert}$ are obtained by projecting group-level parameters $\varphi_{\text{group}} \in \mathbb{R}^{s_{\text{DSR}} \times \vert\varphi\vert}$ through low-dimensional subject embeddings $S^{(j)} \in \mathbb{R}^{s_{\text{DSR}}}$,
\begin{equation}
\varphi^{(j)} = S^{(j)} \varphi_{\text{group}} .
\end{equation}
The shared parameters describe a family of possible dynamics; the embedding $S^{(j)}$ selects the subject associated with subject $j$. Learning them jointly constrains each reconstruction to this common family while allowing systematic differences between systems \citep{ICLR2025_d4c96180}. This is the sense in which we use hierarchy throughout the method.

We fit the reconstruction model and embeddings using the recorded trajectories $\{x_t\}$, before any policy training. We assess whether generated trajectories reproduce the observed state distributions and temporal structure, in addition to prediction accuracy. Sufficient for
this is the agreement of the \gls{kl} divergence $D_{\mathrm{KL}}$ and temporal consistency using the dimension-wise Hellinger distance $D_\mathrm{H}$ between power spectra \citep{10.1371/journal.pcbi.1007263,pmlr-v162-brenner22a,NEURIPS2022_495e55f3,pmlr-v202-hess23a}. Appendix \ref{sec:appendix_training} gives the reconstruction objectives, training procedure, and dataset-specific settings. The resulting model supplies a transition for each subject and an embedding that will also identify that subject to the policy.

\subsection{Turning the reconstruction into a control environment / world model}
The learned reconstruction predicts how a system evolves without intervention. We define an action $a_t\in\mathbb{R}^L$ as a perturbation of the latent state before the learned transition: 
\begin{align}
    z_{t+1}^{(j)}=F_{\varphi^{(j)}}(z_{t}^{(j)} + a_{t}^{(j)}, s_{t}^{(j)}).
\end{align}
Importantly, actions are only nonzero in the latent dimensions that are decoded (\figref{fig:Methods_3}). 
Thus, policy learning can explore the predicted consequences of state perturbations without requiring recorded control actions. Its validity depends on the accuracy of the reconstructed dynamics and on whether the prescribed perturbation represents the intervention of interest.

\paragraph{Restricting where actions act} The allowed perturbations depend on the task. For example, we may want to permit an action to affect neural activity while requiring behavior to change only through the dynamical coupling. Linear observation decoders make such restrictions possible, by projecting policy proposals onto the common null space of the decoder weights for outputs that must remain unchanged by the immediate perturbation:
\begin{align}
    {a'}_t^{(j)}\sim \pi^{(j)}(\cdot\vert z_t^{(j)}), \quad a_t^{(j)}= N^{(j)}{a'}_t^{(j)}.
\end{align}
Here, $N^{(j)}\in\mathbb{R}^{L\times L}$ is a projection onto the desired null space. In the synthetic benchmark, this construction restricts which observed state coordinates can be perturbed. In the neural examples, it prevents the policy from changing the behavioral modality directly. Appendix \ref{sec:appendix_sac} describes the projection implementation. 

\paragraph{Reward function} After applying an action and advancing the model, we decode the next state and evaluate its distance from the target state. Let $D^{(j,m)}$ denote the decoder for modality $m$ of subject $j$,
and let $D_T$ be the set of \emph{target decoders}, i.e., modalities whose decoded outputs are required to reach predefined targets $x^{(m)}_{\mathrm{target}}$. The reward associated with the transition $z_{t}^{(j)} \to z_{t+1}^{(j)}$ is
\begin{align}
r^{(j)}(z_{t+1}^{(j)}, a_{t}^{(j)}) =
&- \sum_{m \in D_T}
\mathop{\mathbb{E}}\limits_{x_{t+1}^{(j, m)}}
\left[
\left\Vert x_{t+1}^{(j,m)} - x^{(m)}_{\mathrm{target}} \right\Vert_2^2
\right]- \lambda_{pen} \left\Vert a_{t}^{(j)} \right\Vert_1,
\label{eq:reward}
\end{align}
where the first term penalizes deviations from target states in the relevant observation modalities, and the second term regularizes control effort. Targets and the action penalty are task-specific hyperparameters. During training, we approximate the decoder expectation with one sample; during evaluation, we use the decoder expected value. The model can now produce a complete transition for \gls{rl}: propose an action, enforce its constraints, advance the latent dynamics, and evaluate the decoded outcome.

\subsection{Learning a policy shared across systems}
We learn the policy using the \gls{sac} algorithm
\citep{pmlr-v80-haarnoja18b}, which maximizes expected discounted reward with an entropy bonus encouraging exploration. \gls{sac} trains a feedforward policy network $\pi_{\psi}(a\mid z)$ and two action-value networks
$Q_{\theta_1}(z,a)$ and $Q_{\theta_2}(z,a)$. The policy maps latent states to action distributions, while the $Q$-networks estimate the return from taking an action and subsequently following the policy, including the entropy bonus.

Training uses transitions $(z_t,a_t,r_t,z_{t+1})$ stored in a replay buffer $\mathcal{D}$. For the purpose of readability, the subject indices will be omitted from now. Each $Q$-network minimizes
\begin{align}
\mathcal{L}_{Q_k}(\theta_k)=\mathbb{E}_{(z_t,a_t,r_t,z_{t+1})\sim\mathcal{D}}
\left[\left(Q_{\theta_k}(z_t,a_t)-y_t\right)^2\right],
\qquad k\in\{1,2\}.
\end{align}
The target $y_t$ combines the immediate reward (\eqref{eq:reward}) with an entropy-adjusted estimate of the cumulative future reward. The policy minimizes
\begin{align}
\mathcal{L}_{\pi}(\psi) = \mathbb{E}_{z_t\sim\mathcal{D},\,a_t\sim\pi_{\psi}(\cdot\mid z_t)}
\left[\alpha\log\pi_{\psi}(a_t\mid z_t)-\min_{k\in\{1,2\}}Q_{\theta_k}(z_t,a_t)\right].
\end{align}

Our extension parameterizes $\pi_\psi$ and $Q_\theta$ through shared trainable group-parameters and the subject embeddings $S^{(j)}$ learned during reconstruction (implicitly assuming that the features which distinguish subjects in their dynamics are also those that matter for control). For instance, the weight matrix of layer $i$ for subject $j$ is
\begin{align}
    W_i^{(j)}=S^{(j)}P_{W_i},
\end{align}
where $P_{W_i}$ is a learned group-level matrix. This construction applies to all weights and biases of both networks. Their respective losses update the shared parameters, while the reconstruction model and embeddings remain fixed.
Sharing parameters restricts each controller's ability to specialize to its reconstructed environment.

Training alternates between collecting generated transitions and updating the networks from the replay buffer. We add Gaussian process noise during trajectory generation to reduce sensitivity to deterministic reconstruction errors. Further \gls{sac} implementation
details are provided in Appendix \ref{sec:appendix_sac}.

\paragraph{Generalization to unseen systems via zero-shot adaptation} To control a dynamical system \emph{unseen} by the policy network, we only fit its subject embedding $S^{(j)}$ using its action-free trajectory data, keeping all shared reconstruction parameters fixed. Inserting $S^{(j)}$ into the fixed policy and Q-networks instantly yields a controller without any policy optimization or fine-tuning. Adaptation therefore uses the new system’s observed dynamics to select a policy from the learned family.

\subsection{Understanding how actions modify the dynamics}
\label{sec:actions}

Beyond generalization, a major advantage of our approach is that the \gls{alrnn} allows us to examine how a trained controller changes the reconstructed dynamics. Its $P$ ReLU units divide state space into $2^P$ \glspl{lsr}, each defined by which units are active. Within each \gls{lsr}, \glspl{fp} and their stability can be computed in closed form, providing direct access to mechanistic properties of the reconstructed system \citep{brenner2026uncovering}. To express this, we collect the activation indicators in a diagonal matrix $D_t$.
Its first $L-P$ diagonal entries are one because these units are linear; the remaining ones satisfy
\begin{align}
d_{t,l}=\mathbf{1}_{\{z_{t,l}>0\}}
\quad \text{for } l>L-P, \qquad
D_t=\mathrm{diag}(1,\ldots,1,d_{t,L-P+1},\ldots,d_{t,L}).
\end{align}
The activation pattern determines the local transition matrix $M_{D_t}=A+WD_t$ such that we can write
\begin{align}
z_{t+1}=M_{D_t}z_t+Cs_t+h.
\label{eq:linearization}
\end{align}
For a constant action $a$, we denote the activation pattern at the perturbed state by $D_t^a$. In the absence of external inputs,
the controlled transition is
\begin{align}
z_{t+1}=M_{D_t^a}(z_t+a)+h.
\label{eq:linearization_actions}
\end{align}
Solving for \gls{fp}s $z_{t+1}=z_t=z^*$ gives
\begin{align}
z^*=(I-M_{D^a_t})^{-1}(M_{D^a_t}a+h),
\label{eq:fp}
\end{align}
provided the inverse exists. This candidate may be \emph{non-virtual}, i.e., the \gls{fp} lies within the same \gls{lsr} in which it is computed, or \emph{virtual}, meaning that the \gls{fp} or the trajectory toward it crosses into another subregion. The effective linear dynamics are thus determined by the occupied \gls{lsr} through $D_{t}^a$, which itself may change under the applied action. As a result, the effect of actions can be interpreted as pushing the system into an \gls{lsr} with distinct dynamical properties and subsequently shifting the \glspl{fp} within this region. We use these properties to examine which regions the controller visits, how close their \gls{fp}s lie to the target, and the eigenvalues of their transition matrices.

\begin{figure}[t]
    \centering
    \captionsetup[subfigure]{position=top, justification=raggedright, singlelinecheck=false}
    \setcounter{subfigure}{0}
    \begin{subfigure}{0.65\linewidth}
        \caption{}
            \label{fig:Methods_1}
        \includegraphics[width=\linewidth]{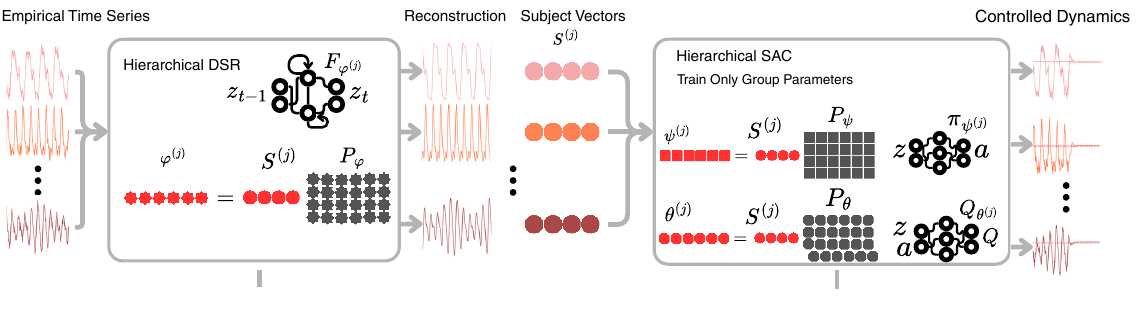}
    \end{subfigure}
    \setcounter{subfigure}{2}
    \begin{subfigure}{0.33\linewidth}
        \caption{}
        \vspace{-20pt}
        \label{fig:Methods_3}
        \includegraphics[width=1.0\linewidth]{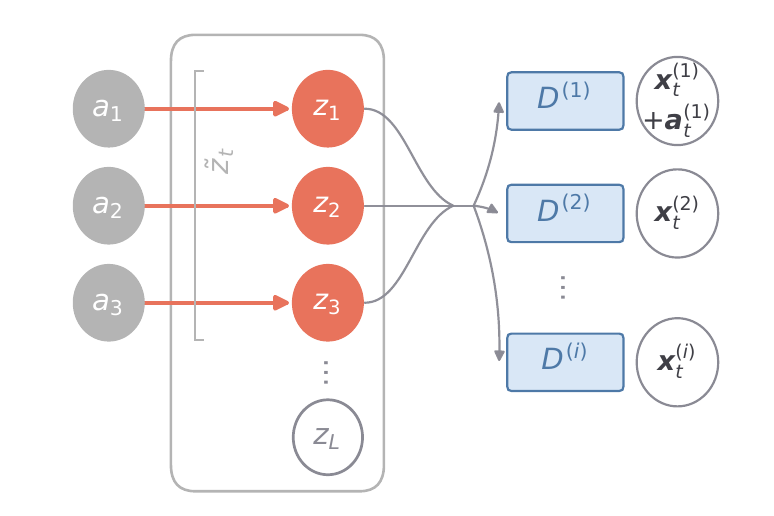}
    \end{subfigure}
    \setcounter{subfigure}{1}
    \begin{subfigure}{0.65\linewidth}
        \vspace{-10pt}
        \caption{}
        \vspace{-8pt}
        \label{fig:Methods_2}
        \includegraphics[width=\linewidth]{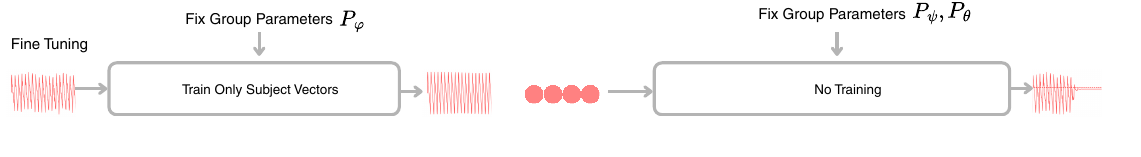}
        \vspace{20pt}
    \end{subfigure}
    \setcounter{subfigure}{3}
    \begin{subfigure}{0.33\linewidth}
        \caption{}
        \vspace{-10pt}
        \includegraphics[width=1.0\linewidth]{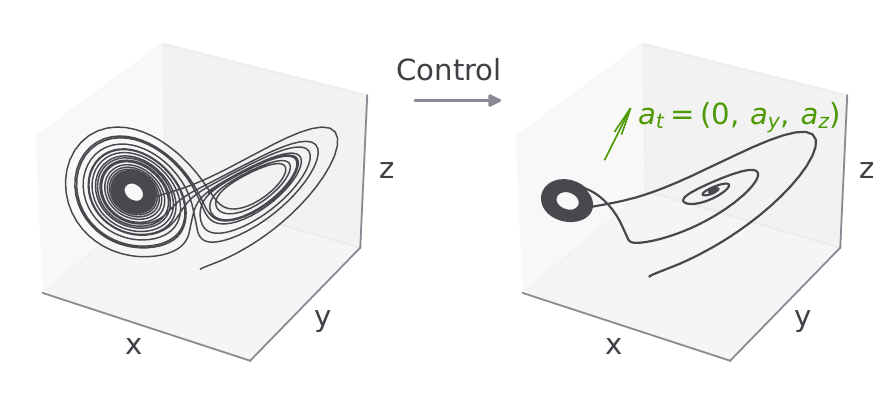}
    \end{subfigure}
    \caption{\textbf{Framework overview.} \textbf{a}: A hierarchical \gls{dsr} model is trained on multiple related dynamical systems (“subjects”) and used as world model for \gls{mbrl}. Subject-specific parameters $\varphi^{(j)}$ are obtained by projecting shared parameters $P_\varphi$ onto low-dimensional subject embeddings $S^{(j)}$.
    These embeddings are reused to condition the policy $\pi_{\psi}$ and value networks $Q_{\theta}$ by modulating group-level parameters $P_{\psi}$ and $P_{\theta}$.
    \textbf{b}: For unseen subjects, learning only a new embedding via \gls{dsr} fine-tuning enables insertion into the control model.
    \textbf{c}: The high-dimensional dynamics are controlled only in the directions that are decoded ($\tilde{z}_t$). To underactuate the problem and for application-specific constraints,
    actions are constrained to be orthogonal to the target modalities and target dimensions.
    \textbf{d}: In the Lorenz-63 benchmark, the actions only affect the $y-$ and $z-$directions.}
\end{figure}

\section{Experiments}
We first applied our approach to Lorenz-63 \citep{DeterministicNonperiodicFlow} and double-pendulum systems, where we could transfer the learned actions directly to the true underlying equations. We then examined whether a trained controller could generalize to unseen  Lorenz systems using only action-free observations. Finally, we used neural and behavioral recordings to test in-silico whether the model could predict actions that suppress movement without directly changing the decoded behavioral variables.

\subsection{Evaluation and comparison methods}
For the synthetic benchmarks, we distinguished performance in the learned (world) model from performance on the true system. To test this transfer, we encoded the current true-system observation, obtained an action from the policy, and mapped it back to the permitted system coordinates. We then applied this perturbation to the true system and advanced its equations numerically.\\
We used three comparisons to examine different parts of the approach. First, \textbf{independent \gls{alrnn}-\gls{sac}} learns a separate policy and value function for each system, using the same reconstructed environments, action constraints, and reward as the hierarchical controller. This comparison tests whether hierarchizing \gls{sac} parameters and reusing subject embeddings improves performance beyond learning each policy independently. Second, \textbf{\gls{alrnn}-CEM-\gls{mpc}} uses the same reconstructed dynamics but chooses actions through online planning \citep{pmlr-v155-pinneri21a}. This comparison asks how a policy learned in simulation performs against repeated online planning. Neither approach uses action-conditioned transitions from the true system for training. Third, we compared learning \emph{without} actions against learning \emph{with} actions, using two established baselines: \textbf{DreamerV3}, which learns an explicit policy through imagined trajectories from controlled interactions \citep{Hafner2025}, and \textbf{TD-\gls{mpc}}, which selects actions through repeated planning in a learned model during deployment \citep{pmlr-v162-hansen22a}. Both baselines were trained separately for each subject.

Within each synthetic benchmark, methods use the same reward, permitted action space, evaluation horizon, and initial-state distribution. We average reward over initializations and time, excluding the first 15 steps, take the median across subjects, and report the mean and standard deviation across training seeds. Reward scales differ between benchmarks, so numerical comparisons are made within each task. Data generation and reconstruction details are provided in Appendix \ref{sec:appendix_training}, with baseline implementations in Appendix \ref{sec:appendix_baseline}.

\subsection{Transfer of control in families of benchmark dynamical systems}

\begin{figure}[ht]
    \centering
    \captionsetup[subfigure]{position=top, justification=raggedright, singlelinecheck=false}
    \begin{subfigure}{0.28\textwidth}
        \centering
        \caption{}
        \vspace{-8pt}
        \includegraphics[width=\linewidth]{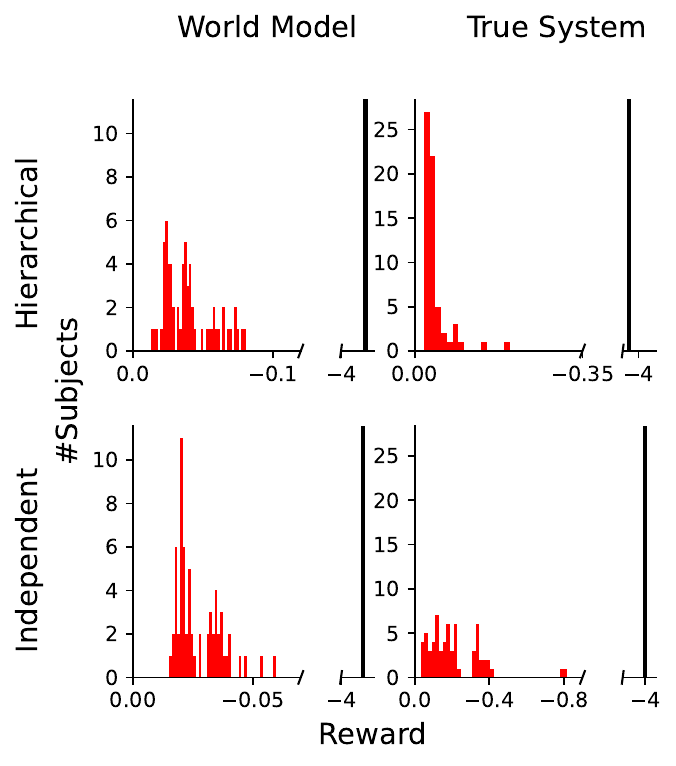}\label{fig:sac_hier_lorenz_reward_comparison_combined_width_43_height_49}
    \end{subfigure}
    \hfill
    \begin{subfigure}{0.355\textwidth}
        \centering
        \caption{}
        \vspace{1pt}
        \includegraphics[width=\linewidth]{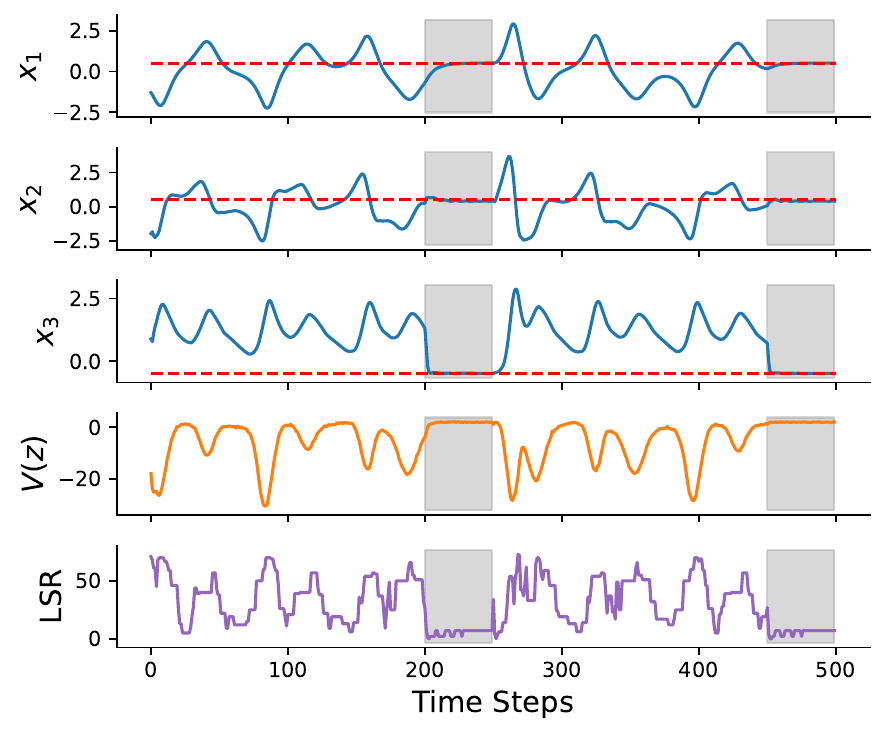}
        \label{fig:sac_hier_sim_lorenz_example_53_width_77_height_63}
        \vspace{-11pt}
    \end{subfigure}
    \begin{subfigure}{0.35\textwidth}
        \centering
        \caption{}
        \vspace{32pt}
        \includegraphics[width=\linewidth]{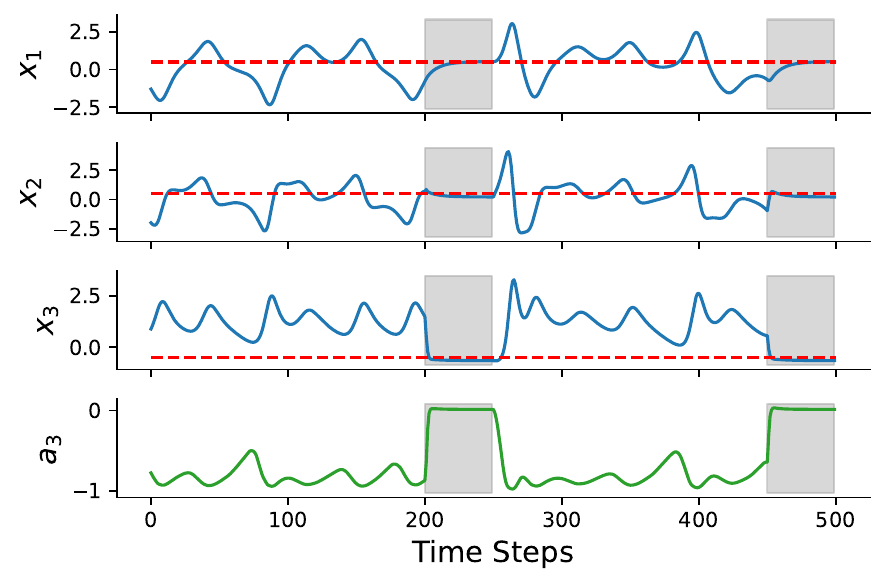}
        \vspace{-12pt}
        \label{fig:sac_hier_true_lorenz_example_53_width_77_height_50}
    \end{subfigure}    
    \begin{subfigure}{0.25\linewidth}
        \caption{}
        \includegraphics[width=\linewidth]{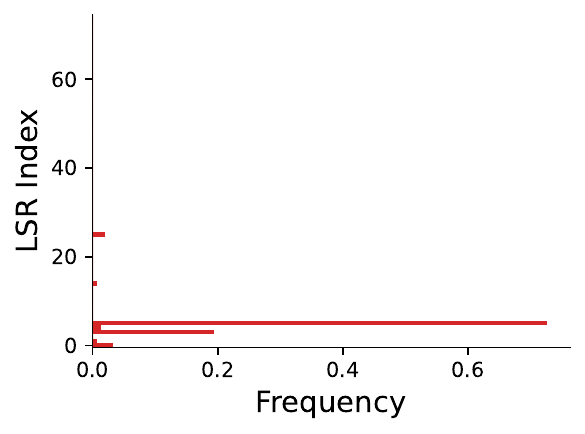}
        \label{fig:sac_hier_sim_lorenz_nlsr_hist_subject_53_width_37_height_27}
    \end{subfigure}
    \begin{subfigure}{0.25\linewidth}
        \caption{}
        \includegraphics[width=\linewidth]{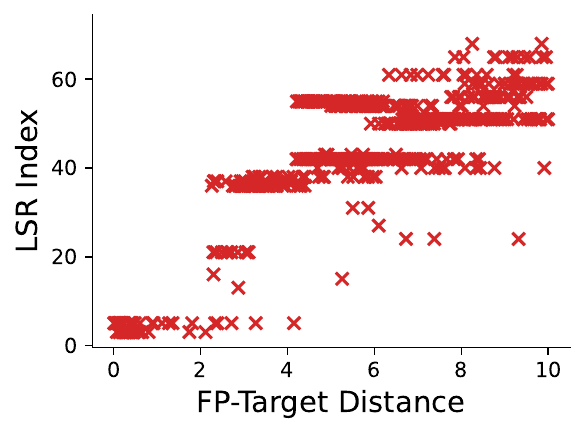}
        \label{fig:sac_hier_sim_lorenz_fp_distance_subject_53_width_37_height_27}
    \end{subfigure}
    \begin{subfigure}{0.25\linewidth}
        \caption{}
        \includegraphics[width=\linewidth]{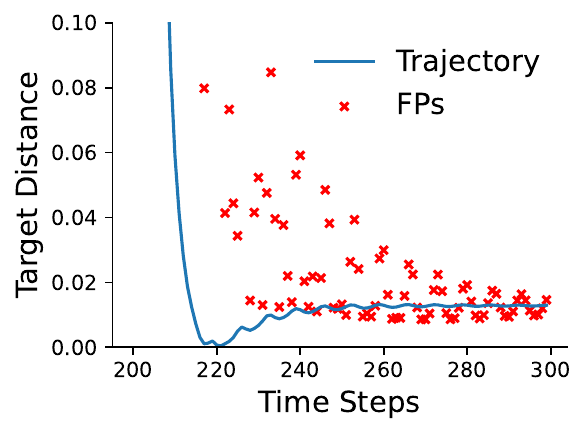}
        \label{fig:sac_hier_sim_lorenz_fp_distance_across_time_subject_53_width_37_height_26}
    \end{subfigure}
    \hspace{100pt}
     \hfill
    \caption{\textbf{Controlling the Lorenz-63 benchmark.} \textbf{a}:
    Reward histograms for randomly initialized trajectories across all subjects for the simulated system (left) and the true system (right) using hierarchically (top) and independently trained models (bottom). Black line denotes uncontrolled baseline.
    \textbf{b}: Representative Lorenz–63 trajectories with reward $=-0.020$ generated by the world model under \gls{sac}
    control during uncontrolled and controlled phases (gray shading), with the value function and occupied \glspl{lsr} shown
    below. While $x_2$ and $x_3$ are directly controlled and rapidly approach the target (dashed red line), $x_1$ converges to a nearby \gls{fp}. 
    \textbf{c}: Corresponding trajectories of the true system, demonstrating sim-to-real transfer, with a reward of 
    $-0.03$.
    \textbf{d, e}: For the same subject, the most frequently occupied subregions (indices 3 and 5) are those admitting non-virtual \glspl{fp} (red crosses) close to the target. \textbf{f}: During extended control, the trajectory approaches the target while action-dependent \glspl{fp} are progressively shifted toward it.}
    \label{fig:fig2}
\end{figure}

\paragraph{Lorenz-63 benchmark} We first considered 64 Lorenz-63 systems with different parameter combinations $(\rho, \sigma, \beta)\in \{36, 66, 96, 126\}\times\{8, 9, 10, 11\}\times\{1, 2, 3, 4\}$ spanning  periodic and chaotic dynamics (\eqref{eq:lorenz}). We treat these as distinct subjects on which we jointly train the hierarchical \gls{dsr} model, obtaining four-dimensional subject-specific subject vectors $S^{(j)}$. Decoders select the first 3 latent dimensions (Appendix \ref{sec:appendix_lorenz}), allowing  latent actions to act as direct increments of the Lorenz-63 coordinates, and thus be interpretable in original physical units and applicable to the true system without inversion (\twofigref{fig:Methods_1}{fig:Methods_3}). The models achieve successful reconstruction, with median \gls{kl} divergence $D_{\mathrm{KL}}=1.8$ and Hellinger distance $D_\mathrm{H}=0.15$ (\twofigref{fig:hier_lorenz_subject_vectors_pca_width_53_height_35}{fig:lorenz_3d_reconstruction_width_157_height_142}).

The control objective was to steer each system toward a target state of $x_{\mathrm{target}} = (0.5, 0.5, -0.5)$, while constraining actions to change only two of its three coordinates and thus underactuating the problem (\figref{fig:Methods_3}; results obtained from controlling only one dimension are found in \figref{fig:sac_hier_true_lorenz_1_dim_example_53_width_57_height_37}).

Independent policies achieved a higher reward ($-0.0232 \pm 4\cdot10^{-4}$) than the hierarchical policy ($-0.047 \pm 0.006$). Allowing each controller to specialize therefore improved performance in the models used for training. However, when we applied the policies to the true Lorenz dynamics (sim-to-real transfer), hierarchical \gls{sac} achieved $-0.040 \pm 0.004$, whereas independent \gls{sac} achieved $-0.171 \pm 0.017$ 
(\twofigref{fig:sac_hier_lorenz_reward_comparison_combined_width_43_height_49}{fig:sac_hier_true_lorenz_example_53_width_77_height_50}). Moreover, hierarchical training reduced the number of trainable parameters by $94\%$ compared to independently trained models. Thus, a single hierarchized policy can successfully control a diverse family of nonlinear dynamics, while parameter sharing avoids overfitting to individual model errors (see \figref{fig:sac_hier_true_lorenz_all_controlled_trajectories_width_197_height_198} for all systems).

The other methods show how this transfer compares with alternative algorithms. CEM-\gls{mpc} achieved $-0.0442\pm 8\cdot10^{-5}$ using the same \gls{alrnn} reconstructions, placing its performance close to that of hierarchical \gls{sac}, but with online planning. TD-\gls{mpc} achieved the highest reward, $-0.0328\pm9\cdot10^{-4}$, and Dreamer achieved $-0.047\pm0.001$, both after learning from controlled interactions. On this benchmark, the hierarchical policy therefore reached a similar performance range \emph{without} observing actions during training and \emph{without} online planning at deployment (Table \ref{tab:reward_results}).

\begin{table}[t]
\centering
\caption{\textbf{Control performance}. The reward is reported with mean and standard deviation across seeds.}
\label{tab:reward_results}
    \begin{small}
        \begin{tabular}{lcccc}
            & \multicolumn{2}{c}{Lorenz-63} & \multicolumn{2}{c}{Double-pendulum} \\
            \cmidrule(lr){2-3}\cmidrule(lr){4-5}
            & World Model & True Environment & World Model & True Environment \\
            \midrule
            \gls{alrnn}-SAC, Hierarchical &$-0.047 \pm 0.006$
            & $\mathbf{-0.040 \pm 0.004}$   & $-47.93\pm 10.57$ & $-9.74\pm2.42$    \\

            \gls{alrnn}-SAC, Independent & $-0.0232 \pm  4\cdot10^{-4}$
            &$-0.171 \pm 0.017$   & $-51.00 \pm 18.47$
            & $-16.82 \pm 7.78$ \\

            \gls{alrnn}-CEM-MPC, Independent &-$0.02090\pm4\cdot10^{-5}$
            &$-0.04424 \pm8\cdot10^{-5}$ & $-15.35 \pm 0.71$  &
            $-6.34 \pm 0.55$   \\
            \midrule
            Online-TD-MPC, Independent & $-40\pm2$
            & $\mathbf{-0.0328 \pm 9\cdot10^{-4}}$ & $-3547 \pm 323$  &
            $-27.94\pm 7.21$ \\

            Online-DreamerV3, Independent & $-0.034 \pm 0.002$
            & $-0.047 \pm 0.001$ & $-6.40\pm0.78$ &
            $\mathbf{-5.87 \pm 0.30}$   \\
            \hline
        \end{tabular}
    \end{small}
\end{table}

\paragraph{Generalization to unseen systems}
We next asked whether the controller could be reused when the underlying dynamics changed. We generated additional 64 Lorenz-63 systems by shifting parameters: $(\rho, \sigma, \beta)\in \{51, 81, 111, 141\}\times\{8.5, 9.5, 10.5, 11.5\}\times\{1.5, 2.5, 3.5, 4.5\}$. For each new system, we fitted only its subject vectors $S^{(j)}$ to action-free observations while keeping the shared dynamics parameters fixed (\figref{fig:Methods_2}).
As a baseline, we reused the embedding of the corresponding unshifted training system, without fitting a new embedding (one of its nearest neighbors on the parameter grid). On the true system the model achieved $-0.036\pm0.002$ using the newly fitted embeddings, and $-0.046\pm 0.005$ using the training embeddings, yielding a significant improvement (paired t-test; $t(9)=6.59, p<0.001$). Thus, the hierarchical policy generalized to unseen systems through newly fitted embeddings, outperforming the unshifted-embedding baseline and achieving control performance comparable to that on training systems.

\begin{figure}[ht]
    \centering
    \captionsetup[subfigure]{position=top, justification=raggedright, singlelinecheck=false}
    \begin{subfigure}{0.38\textwidth}
        \centering
        \caption{}
        \vspace{8pt}
        \label{fig:pendulum_subject_vectors_pca_width_53_height_30}
        \includegraphics[width=\linewidth]{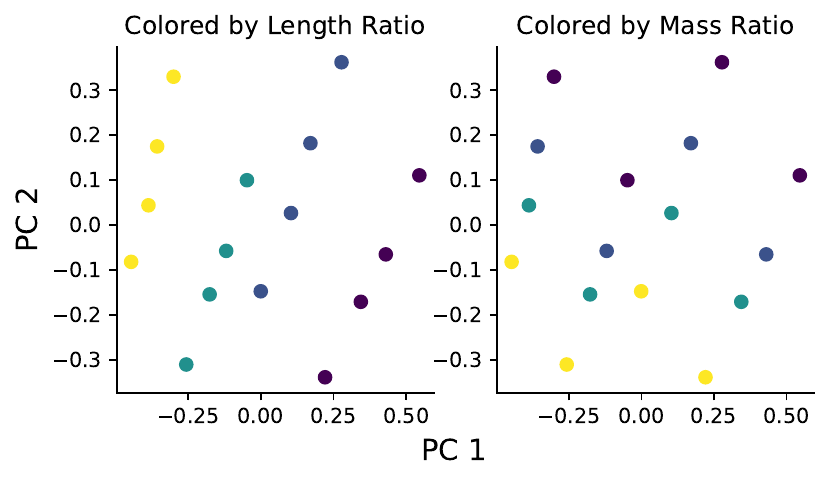}        
    \end{subfigure} 
    \begin{subfigure}{0.59\textwidth}
        \centering
        \caption{}
        \vspace{-8pt}
        \includegraphics[width=\linewidth]{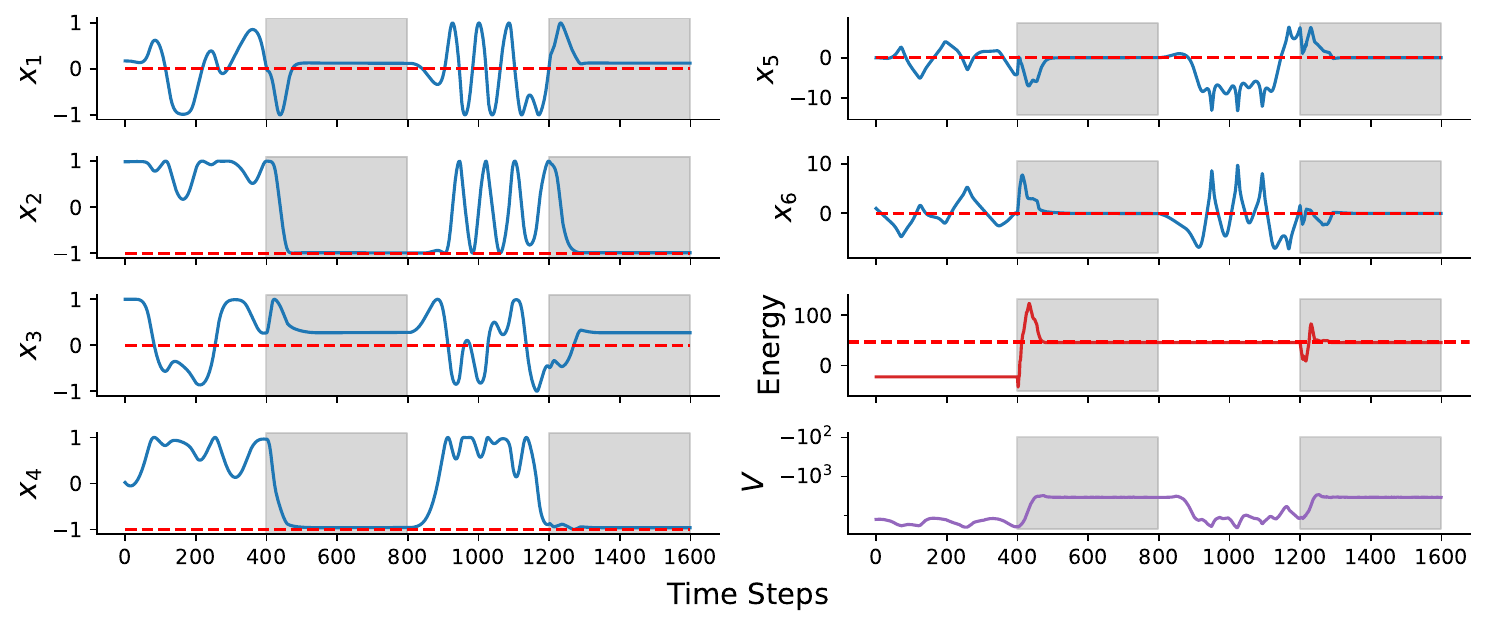}    
        \label{fig:sac_pendulum_example_15_width_97_height_39}
        \vspace{-10pt}
    \end{subfigure} 
    \caption{\textbf{Reconstruction and stabilization of the double-pendulum.}
    \textbf{a}: Colors indicate mass ratios $m_1/m_2$ and length ratios $l_1/l_2$, respectively. Analogous to the Lorenz-63 System, subject vectors cluster according to system parameters.
    \textbf{b}: Example trajectory of the true double-pendulum under control, with the target position indicated by the dashed red line. The 6 variables are sine and cosine of the joint angles as well as the angular velocities.
    Despite its inherent instability the controlled system converges to the target.}
    \label{fig:sac_pendulum}
\end{figure}

\paragraph{Dynamics exploited by the policy}
To gain insights into the mechanisms underlying control, we examined how the policy changed the dynamics within the reconstructed Lorenz-63 systems. Without actions, trajectories visited a wide range of \glspl{lsr} (median of $50$). Under control, in contrast, trajectories concentrated on a small subset of regions. Across subjects, $90\%$ of time steps were spent in at most four 
(\twofigref{fig:sac_hier_sim_lorenz_nlsr_hist_subject_53_width_37_height_27}{fig:sac_hier_sim_lorenz_nlsr_90_summary_width_77_height_18}), although this concentration was not imposed by the reward structure (since the decoder reads out from units that are not subject to the \gls{pwl}
partitioning), and thus rather represents a dynamical preference induced by the control. 
To characterize this preference, we analyzed the local linear dynamics within the occupied regions. For each \gls{lsr}, we compute the \glspl{fp} implied by the linearized controlled system (\eqref{eq:fp}) and measure their distance to the target. We find that the \glspl{lsr} preferentially occupied under control are those whose local dynamics admit \glspl{fp} close to the target (\figref{fig:sac_hier_sim_lorenz_fp_distance_90_summary_width_37_height_27}) and exhibit predominantly stable eigenvalues, indicating saddle-type structure with largely attractive manifolds (\figref{fig:sac_hier_sim_lorenz_eigenvalues_fp_90_summary_width_28_height_27}).
This analysis suggests that the policy directs the system into \glspl{lsr} whose intrinsic dynamics favor convergence (\twofigref{fig:sac_hier_sim_lorenz_nlsr_hist_subject_53_width_37_height_27}{fig:sac_hier_sim_lorenz_fp_distance_subject_53_width_37_height_27}), and shifts the \glspl{fp} ever more closely to the target over time (\figref{fig:sac_hier_sim_lorenz_fp_distance_across_time_subject_53_width_37_height_26}).

\paragraph{Double-pendulum benchmark} 
We further evaluated the framework on a challenging family of 16 double-pendulum systems with varying masses and lengths. The state of this system consists of two angles and their angular velocities. Without damping or control, the dynamics are Hamiltonian and conserve energy, while exhibiting both regular and highly chaotic motion. To ensure continuity, we represented each angle by its sine and cosine and retained both angular velocities, yielding $O=6$ observation dimensions. 

The models once more achieved low median \gls{kl} divergence $D_{\mathrm{KL}}=6.40$ and Hellinger distance $D_\mathrm{H}=0.22$. \Gls{pca} projections of the subject embeddings revealed structure associated with mass and length ratios, which the world model extracted solely from the action-free time series (\figref{fig:pendulum_subject_vectors_pca_width_53_height_30}). This structure emerged although the equations of motion depend nonlinearly on these ratios.

The control objective was to stabilize the double-pendulum in the upright position with both arms pointing up, an unstable \gls{fp} of the uncontrolled dynamics. Actions were constrained to affect only the angular velocities. Hierarchical \gls{sac} again transferred better than independent \gls{sac} ($-9.74\pm2.42$ vs. $-16.82\pm7.78$) and stabilized the true pendulum in $49$ 
percent of the trials on average, from random initial states, trained on action-free recordings alone (\figref{fig:sac_pendulum_example_15_width_97_height_39}). CEM-\gls{mpc} achieved higher rewards using the same reconstructions, suggesting a benefit of online planning on this task. Results for methods trained through interaction with the true system were mixed: DreamerV3 outperformed hierarchical \gls{sac}, whereas TD-\gls{mpc} performed worse (Table \ref{tab:reward_results}).

In contrast to Lorenz-63, all methods performed better on the true systems than in the world models. This suggests that the policies transfer successfully despite imperfect long-horizon world-model fidelity. A likely explanation is that the true system supplies physically consistent states at every step, whereas free-running world-model predictions accumulate model error that lowers the reward attainable in simulation.

On unseen double-pendulum systems, controllers using newly fitted embeddings did not significantly differ in performance from controllers using the embeddings of the nearest training systems (paired $t(9)=-1.30, p=0.208$). The shared policy thus maintained its performance on systems absent from policy training, although fitting system-specific embeddings yielded no additional benefit, unlike on Lorenz-63.

\begin{figure}[ht]
    \centering
    \captionsetup[subfigure]{position=top, justification=raggedright, singlelinecheck=false}
    \begin{subfigure}{0.16\textwidth}
            \centering
            \caption{}
            \vspace{19pt}
            \label{fig:sac_mc_multitask_subject_vectors_corr_width_22_height_35}
            \includegraphics[width=\linewidth]{
                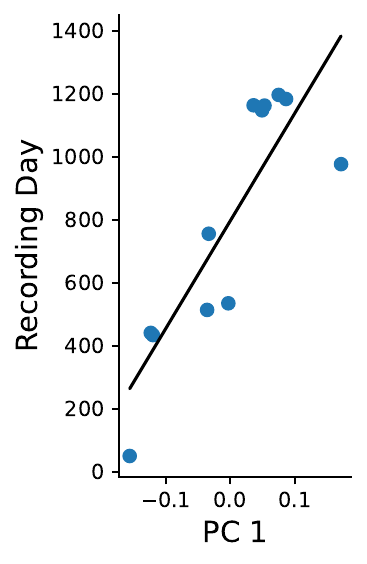
            }
        \end{subfigure}
        \begin{subfigure}{0.34\textwidth}
            \centering
            \caption{}
            \vspace{-8pt}
            \label{fig:sac_mc_multitask_example_trajectories_width_55_height_49}
            \includegraphics[width=\linewidth]{
                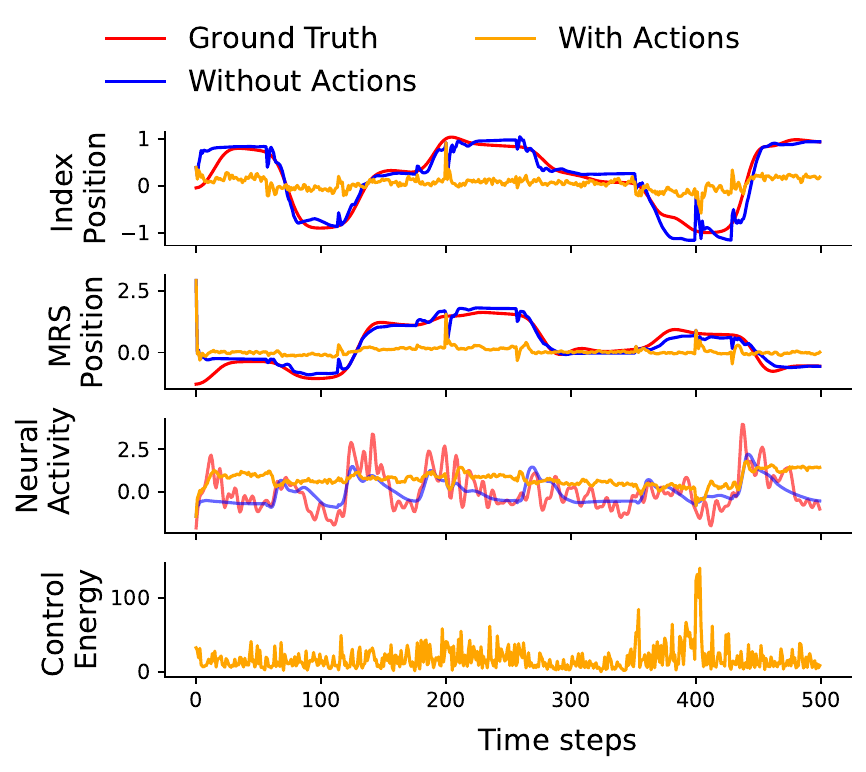
            }
        \end{subfigure}
        \begin{subfigure}{0.34\textwidth}
            \centering
            \caption{}
            \vspace{9pt}
            \label{fig:sac_mc_multitask_action_drop_out_analysis_width_57_height_45}
            \includegraphics[width=\linewidth]{                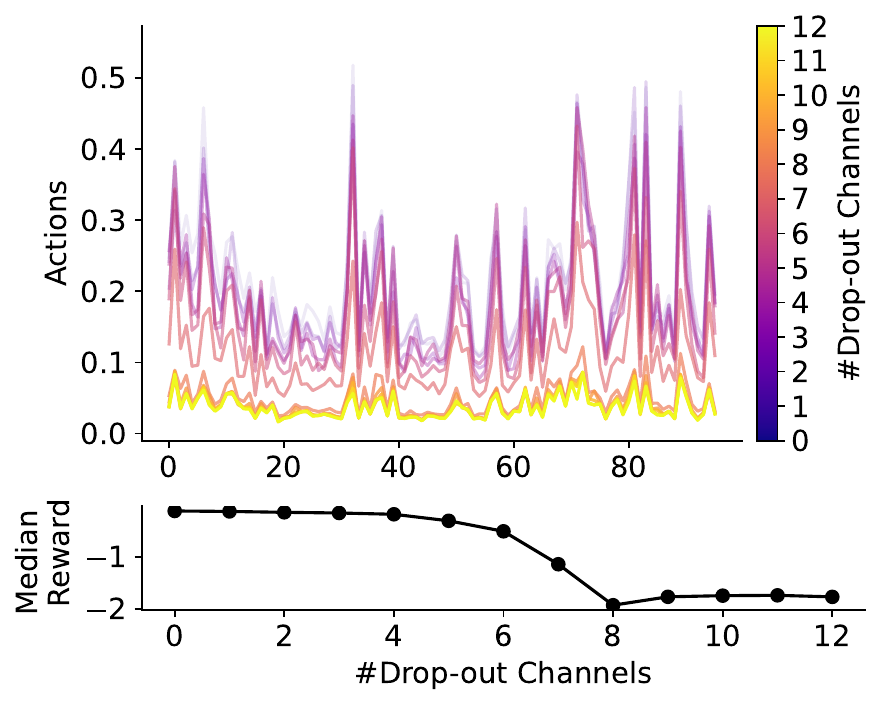
            }
        \end{subfigure}
    \caption{
        \textbf{Reconstruction and model-based control of the LINK data.}
        \textbf{a}: The first principal component of the subject vectors shows a linear relationship with recording days, indicating that the learned subject representations capture structure across sessions.
        \textbf{b}: Finger positions of the index and \gls{mrs} finger groups, neural activity, and control energy for one representative session. 
        Each new goal position, presented to the monkey and provided to the model as external input, triggers a movement in the recorded (red) and simulated (blue) trajectories.
        Under the learned policy (orange), the simulated finger positions stay at rest. Neural activity follows a quasi-periodic dynamics in the uncontrolled model and changes under control.
        Control energy is the squared norm of the decoded action. 
        Controlled trajectories are reinitialized from the recorded state every 200 time steps, which produces the visible jumps.
        \textbf{c}: Action magnitude over time for 0 to 12 excluded channels (color, top) and median reward as a function of the number of excluded channels (bottom). Actions are forced to zero on randomly selected channels. Action magnitude first increases to compensate and is then suppressed entirely once 8 channels are excluded, causing a drop in reward.}
        
\end{figure}

\subsection{Control in models of neural and behavioral recordings}
Unlike the synthetic benchmarks, where policies can be tested directly on the governing equations, validating control predictions in neuroscience requires additional intervention experiments that may be technically or ethically difficult. For the neural datasets, we therefore tested whether policies could suppress \emph{predicted} rather than true movement.
For evaluation, we initialized trajectories from held-out observations and compared them with and without policy actions under the same recorded task conditions. In both datasets, actions were projected to prevent immediate changes in the targeted behavioral output.

\paragraph{Control across recording sessions}
We first used 12 sessions from the longitudinal LINK dataset \citep{dandi:DANDI:001201_0.251023.2336}, which contains neural recordings and finger kinematics from a macaque performing a center-out-reach task, over a period of more than three years. Spiking-band power was measured from 96 channels of Utah arrays in placed motor cortex (see Appendix \ref{sec:appendix_link_data} for more details on the dataset). In this task, the macaque makes finger movements to match visually presented target positions. We treated each session as a subject and learned a shared reconstruction with four-dimensional subject vectors. The embeddings correlated with recording day ($r=0.85, \ p<0.001$; \figref{fig:sac_mc_multitask_subject_vectors_corr_width_22_height_35}), suggesting that they captured systematic changes across recording time. 

The policy was trained to suppress finger movement toward the instructed goal position.
Without actions, the model successfully reproduced finger movements in held-out recordings (\figref{fig:sac_mc_multitask_example_trajectories_width_55_height_49} and Appendix \ref{sec:appendix_link_training}). With actions, it predicted reduced movement across sessions, achieving a median reward of $-0.090\pm 0.002$ (\twofigref{fig:sac_mc_multitask_example_trajectories_width_55_height_49}{fig:sac_mc_multitask_example_trajectories_combined_width_112_height_86}). The comparison shows that, in simulation, the shared controller can suppress predicted behavior across session-specific models while respecting the behavioral constraint. Because actions reach the recorded channels through the decoder, the model also determines how many independent directions are available for intervention. Only 10 of the latent dimensions are decoded, so all 96 channels lie in a 10-dimensional subspace, and two of its directions are reserved for the protected finger outputs. Each channel on which the action is forced to zero removes one of the remaining eight directions. Accordingly, action magnitude first increased as randomly selected channels were excluded, and control failed once eight channels were excluded (\figref{fig:sac_mc_multitask_action_drop_out_analysis_width_57_height_45}). The rank of the decoder and the set of protected outputs thus determine in advance how many constraints on the intervention the system tolerates.

\begin{figure}[ht]
    \centering
    \captionsetup[subfigure]{position=top, justification=raggedright, singlelinecheck=false}
    \begin{subfigure}{0.65\textwidth}
        \centering
        \caption{}
        \vspace{-5pt}
        \includegraphics[width=\linewidth]{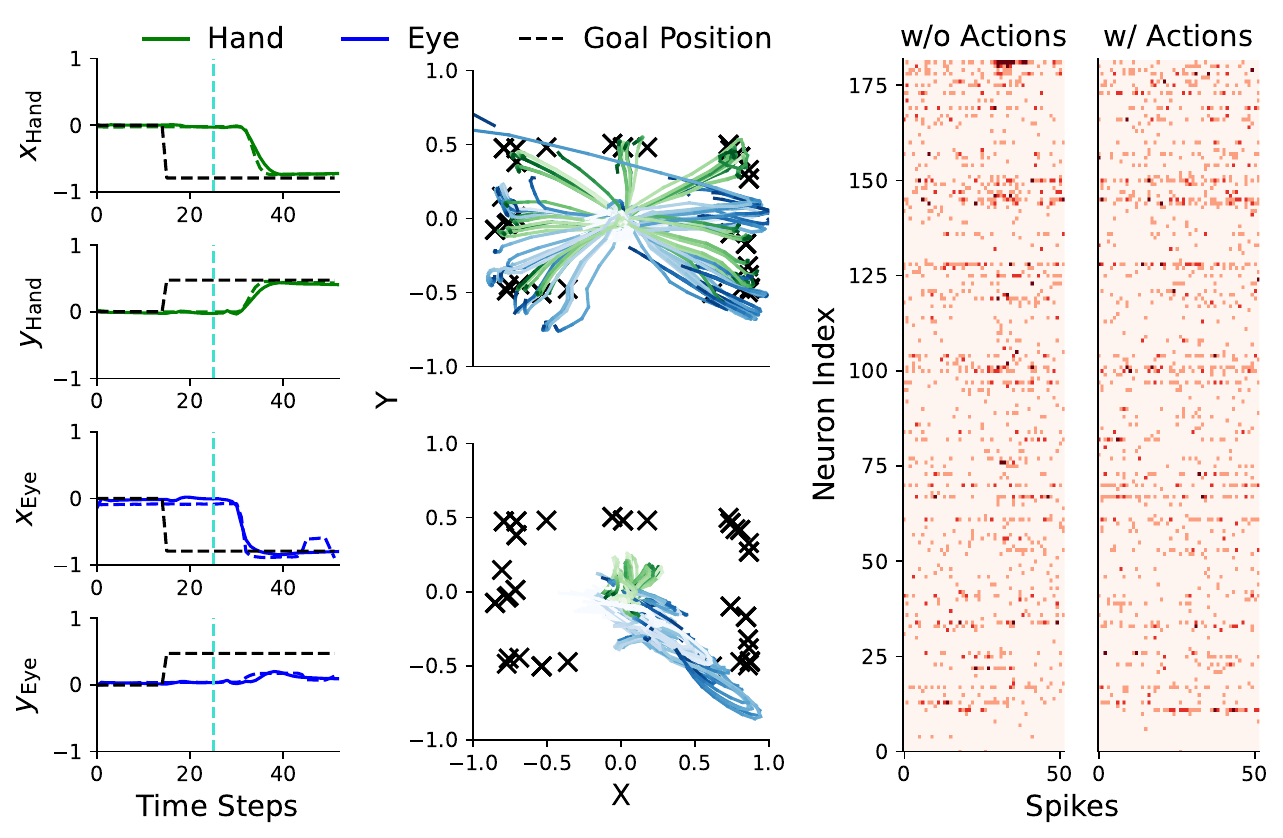} 
        \label{fig:mc_maze_sparse_sac_example_trajectories_actions_vs_no_actions_width_83_height_53}
    \end{subfigure}
    \hfill
    \begin{subfigure}{0.34\textwidth}
        \centering
        \caption{}
        \vspace{8pt}
        \includegraphics[width=\linewidth]{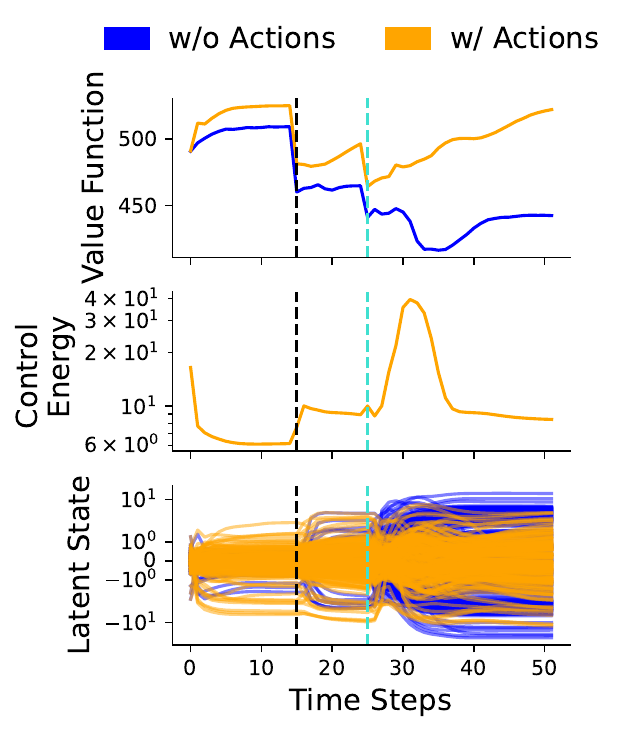} \label{fig:mc_maze_sparse_sac_example_value_function_actions_vs_no_actions_width_39_height_46}
    \vspace{2pt}
    \end{subfigure}    
    \caption{\textbf{Behavioral control through neural stimulation.} 
    \textbf{a, left:}
    Predicted trajectories without control. The model receives the goal position cue (black dashed) and go cue (cyan) as external inputs and generates trajectories (solid) that reproduce the empirical test data (dashed). Hand (green) and eye (blue) positions are shown separately. In successful trials, the hand follows the instructed target, whereas eye position is not consistently target-aligned in either the model or the data.    
    \textbf{center:}
    Two-dimensional trajectories without (top) and with (bottom) control. The policy suppresses hand movement and, to a lesser extent, eye movement despite external inputs driving both toward the target. Color gradients indicate time.
    \textbf{right:} 
    Neural stimulation remains sparse and selective, with no statistically significant changes in mean firing rate  ($t=-1.33$, $p=0.18$) or Fano factor ($t=1.04$, $p=0.30$).
    \textbf{b:}
    Value function, control energy, and latent states for a representative trial. Actions become effective once the goal-position cue is presented and steer the system into a high-value region of latent space associated with trajectories that remain insensitive to the subsequent go cue.
    }
    \label{fig:mc_maze}
\end{figure}
\vspace{-5pt}

\paragraph{Control during delayed reaching} 
To corroborate our findings, we evaluated the modality constraint on a second independent single session from a macaque on a \emph{delayed} center-out reach task \citep{PeiYe2021NeuralLatents}, with spike recordings from 182 motor-cortical neurons together with hand and eye position (more details in Appendix \ref{sec:appendix_mc}). As before, the macaque reached toward a target cue, but now cue presentation was followed by a delay period during which the reach had to be withheld until a go cue appeared. The model received the target location ($x$ and $y$ position) and the go cue as external inputs. Actions once more operated only on the neural dynamics and the policy was trained to suppress the hand movement in spite of the go cue, holding the hand position at $(0, 0)$.

Without actions, on held-out trials, the model generated a reach following the cue (see  \figref{fig:mc_maze_sparse_sac_example_trajectories_actions_vs_no_actions_width_83_height_53} and Appendix \ref{sec:appendix_mc_maze_training} for reconstruction metrics). With actions, it suppressed predicted hand movement despite the task inputs, achieving a reward of $-0.012\pm 0.003$ (\figref{fig:mc_maze}). Interestingly, although the control objective concerned only hand position and actions were restricted to the neural state, predicted eye trajectories also changed, appearing more spatially concentrated under control while retaining substantial movement (\figref{fig:mc_maze_sparse_sac_example_trajectories_actions_vs_no_actions_width_83_height_53}, middle). Furthermore, neural dynamics remained in a physiologically plausible range (\figref{fig:mc_maze_sparse_sac_example_trajectories_actions_vs_no_actions_width_83_height_53}, right).
\Figref{fig:mc_maze_sparse_sac_example_value_function_actions_vs_no_actions_width_39_height_46} illustrates the effect of actions on the latent dynamics and the value function. Only sparse actions are required to suppress movement: once the goal cue is presented, actions prevent the system from being particularly sensitive to the go cue. \Figref{fig:mc_maze_control_mechanism} analyzes the controlled dynamics in terms of \glspl{lsr}.

\section{Discussion}
This work presents a hierarchical \gls{mbrl} framework for learning control policies from action-free recordings, without controlled trajectories or direct interaction with the underlying systems. By combining hierarchical \gls{dsr} with subject-conditioned \gls{sac} policies, the framework shares structure across related systems while preserving subject-specific variation through low-dimensional embeddings.
The \gls{pwl} world model further permits mechanistic analysis of control through occupied \glspl{lsr}, induced \glspl{fp}, and local stability properties, extending previous uses of \gls{pwl} recurrent models for interpreting nonlinear dynamics \citep{Sussillo2013,Linderman2017,smith2021reverse}.

The Lorenz-63 and double-pendulum experiments demonstrate that policies trained strictly on uncontrolled, autonomous trajectories, within a hierarchical framework, can transfer to true systems.
In both benchmarks, the hierarchical policy transferred more reliably than independently trained policies, although these performed as well or better in simulation. This suggests that parameter sharing reduces overfitting to subject-specific world-model inaccuracies.

Beyond the gains in data efficiency and generalization known from hierarchical and multitask RL \citep{NIPS1992_d14220ee, Barto2003,10.5555/3298483.3298491}, parameter sharing plays an additional role here. Since action-free recordings provide no empirical action distribution to constrain the policy \citep{NEURIPS2020_a322852c,NEURIPS2021_f29a1797}, the reduced capacity of the hierarchical policy may act as an implicit regularizer that limits specialization to inaccuracies of individual world models.

The results on unseen Lorenz systems, that is, subject embeddings the controller was never trained on, further indicate that hierarchical training can help controllers generalize, and that learned embeddings capture dynamical variation relevant to control. On unseen double-pendulum systems, the shared policy maintained its performance, although newly fitted embeddings yielded no gain over those of the nearest training systems. The benefit of embedding-based adaptation may therefore depend on how densely the training systems sample the family of dynamics. Repeated online planning partly improved performance, which makes combining the hierarchical policy with planning at deployment a promising extension.

The fact that the policy and value function remain fixed while only the subject-specific representation is adapted indicates that the learned embedding provides a common coordinate system in which both \gls{dsr} and control can be performed, in line with prior work on transferring control policies on learned representations of system dynamics \citep{yu2017preparingunknownlearninguniversal, Killian2017-ds}. The \gls{pwl} structure also clarifies how the policies achieve control. Under control, trajectories concentrate within a small subset of the \glspl{lsr} visited during autonomous dynamics. These target regions harbor \glspl{fp} near the goal and exhibit locally attractive dynamics, revealing the mechanistic basis of policy behavior far better than reward values alone.

In sum, beyond standard closed-loop neural control \citep{Chandrabhatla2023-xm,Horn2025,Krauss2021,https://doi.org/10.1002/hbm.70600}, our framework handles multimodal systems with mismatched intervention and target modalities. By projecting latent actions onto the behavioral decoder's null space, we prevent direct behavioral manipulation, forcing changes to emerge entirely through learned neural–behavioral couplings. While learned policies successfully modify predicted hand trajectories with minimal shifts in neural summary statistics, these results reflect world-model simulations. They should be interpreted as proof-of-concept for model-based feasibility rather than validated physiological control.

\paragraph{Limitations}
Four points limit the generality of our approach. First, the hierarchical parameterization places subject-specific dynamics in linear subspaces, which regularizes the model but may be too restrictive when systems differ nonlinearly. Second, reusing reconstruction embeddings for control assumes that the factors distinguishing subjects' dynamics also determine appropriate control. Although supported by the present experiments, this correspondence need not hold in other settings \citep{pmlr-v120-lambert20a}. Third, adaptation to unseen systems likely requires sufficient coverage of the system family and reconstruction performance.
Finally, representing actions as additive latent perturbations permits decoder-based constraints when no actuator data are available, but leaves open how physical interventions couple to the true system, which distinguishes this work from most \gls{mbrl} frameworks \citep{Hafner2025, moerland2022modelbasedreinforcementlearningsurvey}.

\subsubsection*{Acknowledgments and Funding Disclosure}
This work was funded by the Hector II foundation, by the German Research Foundation within the Research Unit FOR 5159  (subproject 11), by the Wellcome Trust  (FUTURE-D, Z334349/Z/25/Z), by the Federal Ministry of Research, Technology and Space under the neuroAI initiative (NAILIt, grant number 01GQ2509B), and by the Excellence Strategy of the German Federal and State Governments.
\FloatBarrier

\bibliography{hier_control}
\bibliographystyle{tmlr}

\appendix
\newpage
\section{Additional Experimental Results and Analyses}
\subsection{Lorenz-63}
\begin{figure}[!htb]
    \centering
    \includegraphics[width=0.4\linewidth]{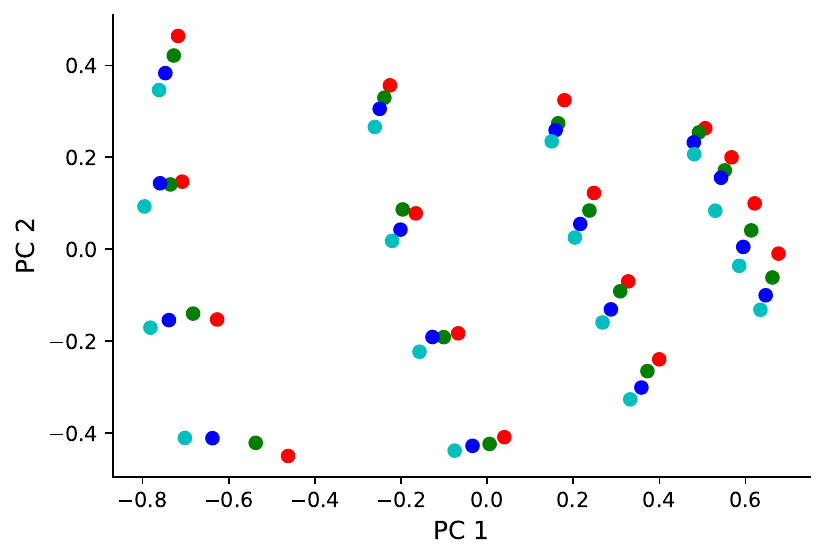}
    \caption{\textbf{Subject vectors cluster according to Lorenz-63 parameters.} \Gls{pca} projection of the subject vectors. Colors indicate constant values of $\sigma$, while vertical and horizontal separation reflect $\beta$ and $\rho$. The subject vectors condition \gls{sac} training and convey information about subject-specific dynamics.}
    \label{fig:hier_lorenz_subject_vectors_pca_width_53_height_35}
\end{figure}

\begin{figure}[!htb]
    \centering
    \includegraphics[width=0.5\linewidth]{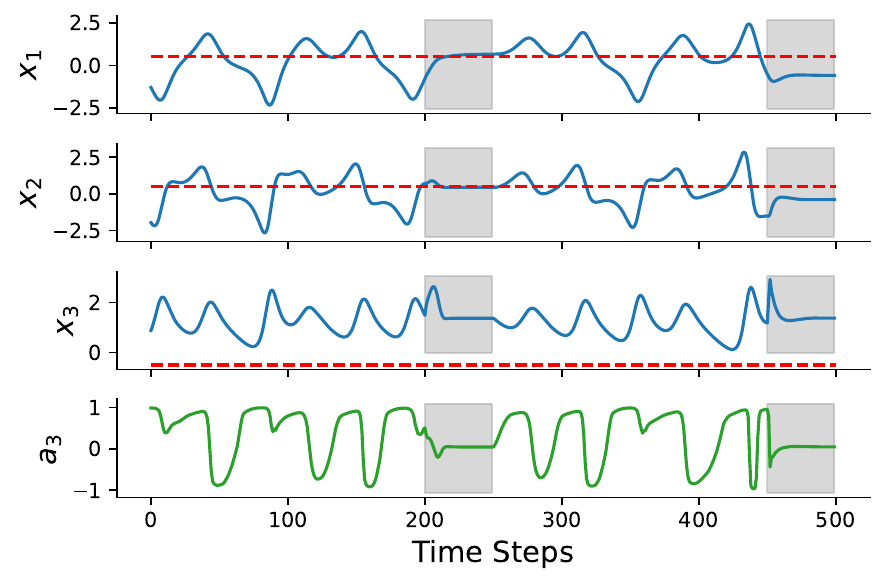}
    \caption{\textbf{Limited controllability under single dimension actuation.} Example trajectory of the true Lorenz-63 system with actions applied exclusively to the third state dimension. The dashed line indicates the target state.}
    \label{fig:sac_hier_true_lorenz_1_dim_example_53_width_57_height_37}
\end{figure}

\begin{figure}[!htb]
    \centering
    \captionsetup[subfigure]{position=top, justification=raggedright, singlelinecheck=false}
    \begin{subfigure}{0.69\linewidth}
        \caption{}
        \includegraphics[width=\linewidth]{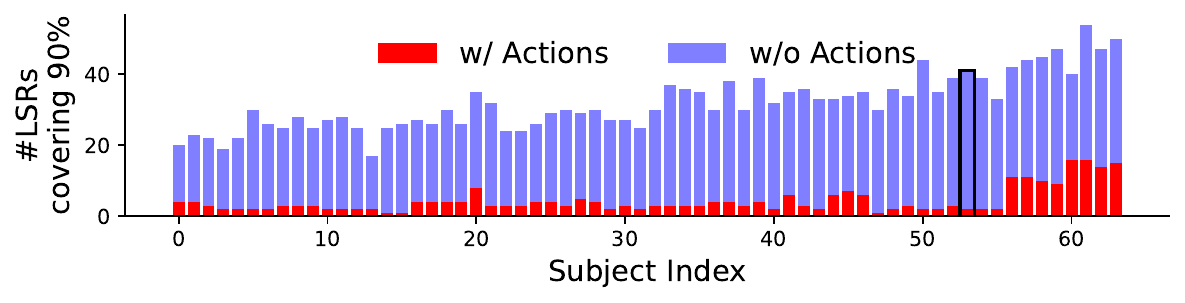}
        \label{fig:sac_hier_sim_lorenz_nlsr_90_summary_width_77_height_18}
    \end{subfigure}
    \begin{subfigure}{0.3\linewidth}
        \caption{}
        \includegraphics[width=\linewidth]{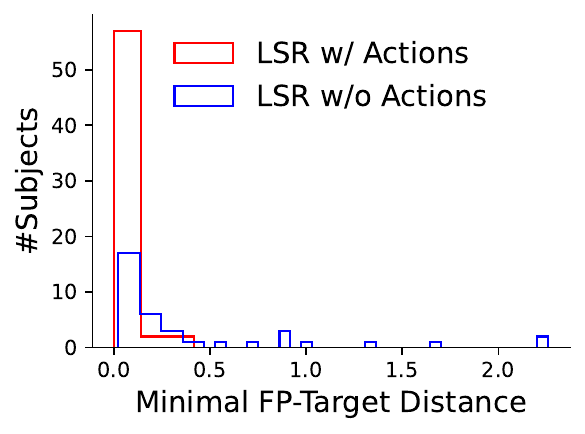}
        \label{fig:sac_hier_sim_lorenz_fp_distance_90_summary_width_37_height_27}
    \end{subfigure}
    \begin{subfigure}{0.30\linewidth}
        \caption{}
        \includegraphics[width=\linewidth]{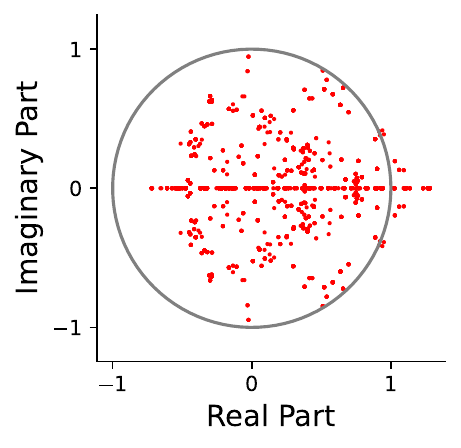}
        \label{fig:sac_hier_sim_lorenz_eigenvalues_fp_90_summary_width_28_height_27}
    \end{subfigure}
    \caption{\textbf{Mechanisms underlying Lorenz-63 control.} \textbf{a}: Number of \glspl{lsr} required to cover 90\% of time steps for uncontrolled (blue) and controlled (red) trajectories of each subject. Control strongly reduces region occupancy. Subject 54 (cf. \figref{fig:fig2}) is highlighted. \textbf{b}: Minimal distance to the target of the non-virtual \glspl{fp} computed from sampled actions in each occupied \gls{lsr} (cf. \figref{fig:sac_hier_sim_lorenz_fp_distance_subject_53_width_37_height_27}). \textbf{c}: Eigenvalues of the Jacobian at the \glspl{fp}, which depends only on the \gls{lsr}, indicate stability along most directions. The unit circle is shown in gray.}
\end{figure}

\begin{figure}[!htb]
    \centering
    \includegraphics[width=\linewidth]{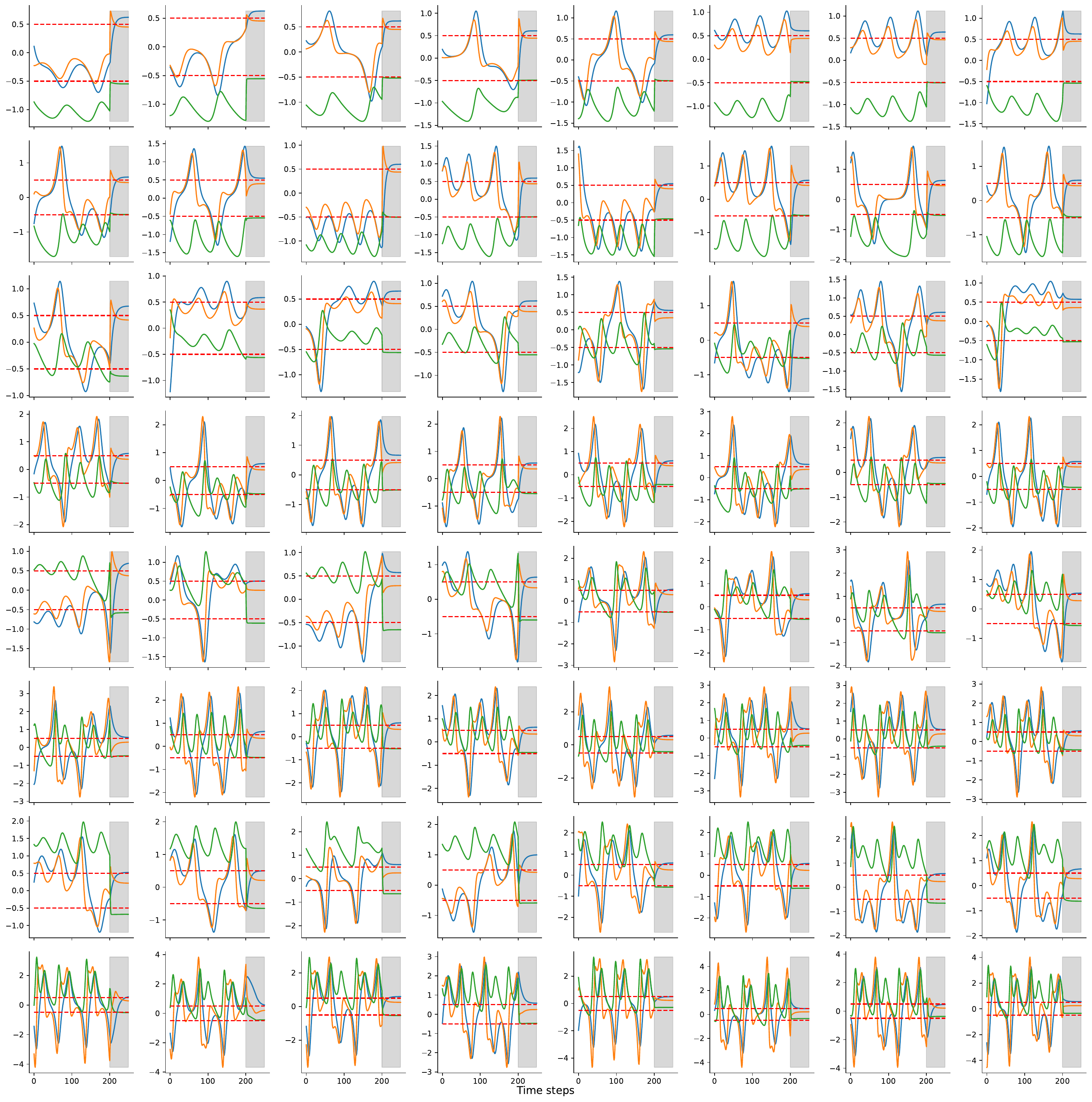}
    \caption{\textbf{Robust control of heterogeneous nonlinear dynamics.} Trajectories of all 64 Lorenz-63 systems under the learned policy. Dashed lines indicate the target state, and gray shading marks the action phase.}
    \label{fig:sac_hier_true_lorenz_all_controlled_trajectories_width_197_height_198}
\end{figure}

\begin{figure}
    \centering
    \captionsetup[subfigure]{position=top, justification=raggedright, singlelinecheck=false}
    \begin{subfigure}{0.49\textwidth}
        \centering
        \caption{}
        \includegraphics[width=\linewidth]{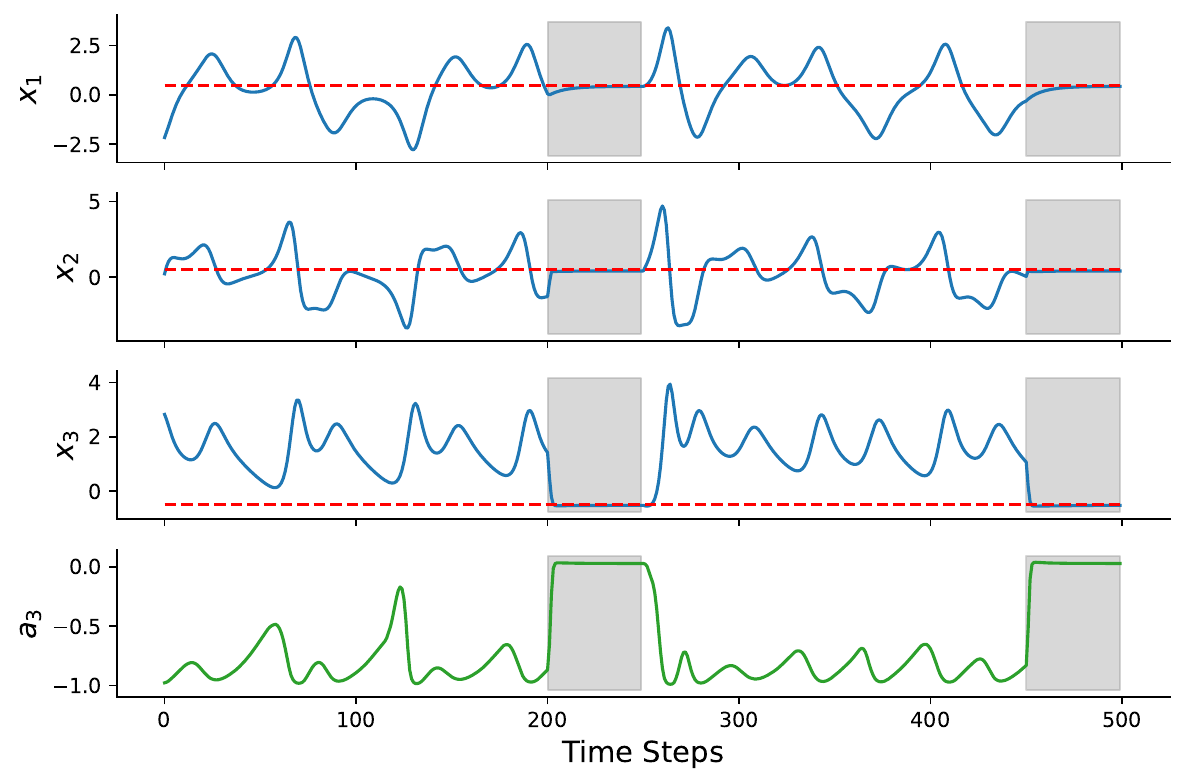}
    \end{subfigure}
    \begin{subfigure}{0.42\textwidth}
        \centering
        \caption{}
        \vspace{22pt}
        \includegraphics[width=\linewidth]{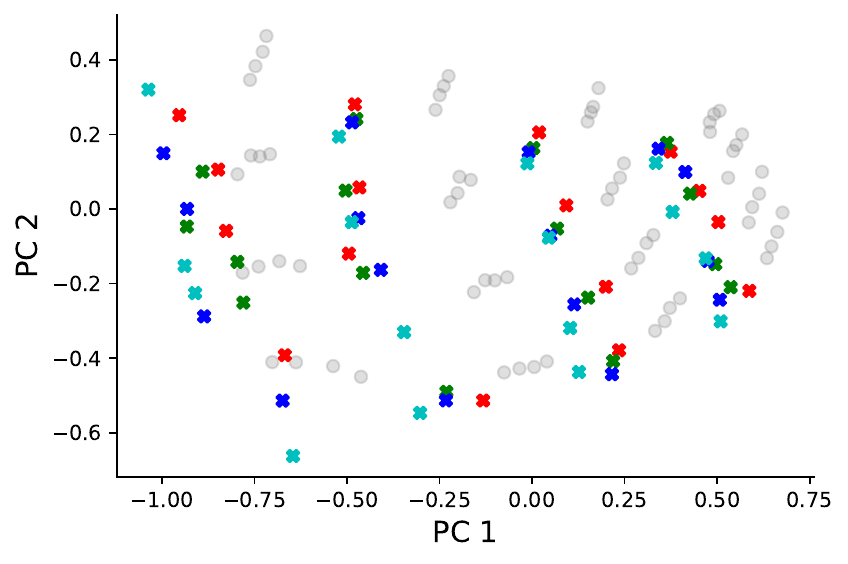}
    \end{subfigure}
        
    \caption{\textbf{Generalization through fine-tuned subject vectors.} On unseen subjects, fine-tuned subject vectors achieved a reward of $=-0.036\pm 0.002$, compared with $-0.046\pm 0.005$ for the unchanged vectors of the nearest training subjects. \textbf{a}: Fine-tuned subject vectors inserted into the fixed \gls{sac} model control the true, unseen systems without further optimization (cf. \figref{fig:sac_hier_true_lorenz_example_53_width_77_height_50}).
    \textbf{b}: \Gls{pca} projection of the subject vectors of the fine-tuned (colored) and original Lorenz-63 systems. Shifted parameter regimes produce corresponding shifts in the \gls{pca} embedding (cf. \figref{fig:hier_lorenz_subject_vectors_pca_width_53_height_35}).}
    \label{fig:fine_tuning}
\end{figure}

\FloatBarrier
\subsection{Macaque delayed center-out reach data}
\begin{figure}[!htb]
    \centering
    \captionsetup[subfigure]{position=top, justification=raggedright, singlelinecheck=false}
    \begin{subfigure}{0.33\linewidth}
        \caption{}
        \includegraphics[width=\linewidth]{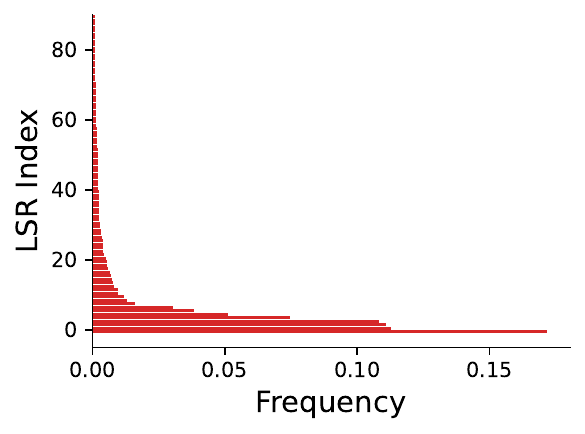}
        \label{fig:sac_maze_hist_lsr_width_37_height_27}
    \end{subfigure}
    \begin{subfigure}{0.33\linewidth}
        \caption{}
        \includegraphics[width=\linewidth]{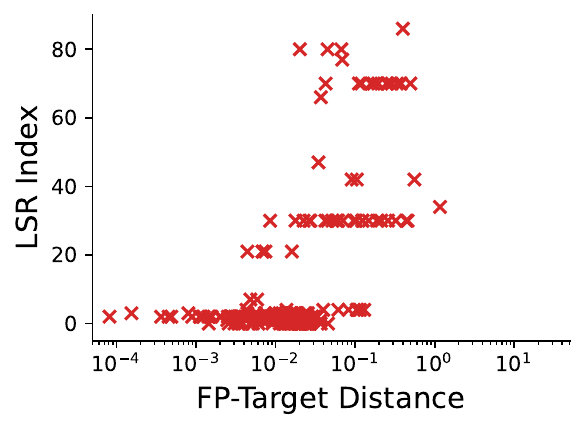}
        \label{fig:sac_maze_fp_distance_width_37_height_27}
    \end{subfigure}
    \caption{\textbf{Control mechanisms in multimodal neural data.}  \textbf{a}: Occupancy of \glspl{lsr} during control. \textbf{b:} Minimal distance of non-virtual \glspl{fp} to the target in each occupied \gls{lsr}. The regions occupied during control are those that admit a non-virtual \gls{fp} close to the target,
    indicating that the policy steers the system into regions of latent space whose linear dynamics favor reaching the goal (cf. \figref{fig:sac_hier_sim_lorenz_nlsr_hist_subject_53_width_37_height_27}-
    \ref{fig:sac_hier_sim_lorenz_fp_distance_subject_53_width_37_height_27}).}
    \label{fig:mc_maze_control_mechanism}
\end{figure}

\FloatBarrier
\subsection{LINK Data}

\begin{figure}[!htb]
    \centering
    \includegraphics[width=0.9\linewidth]{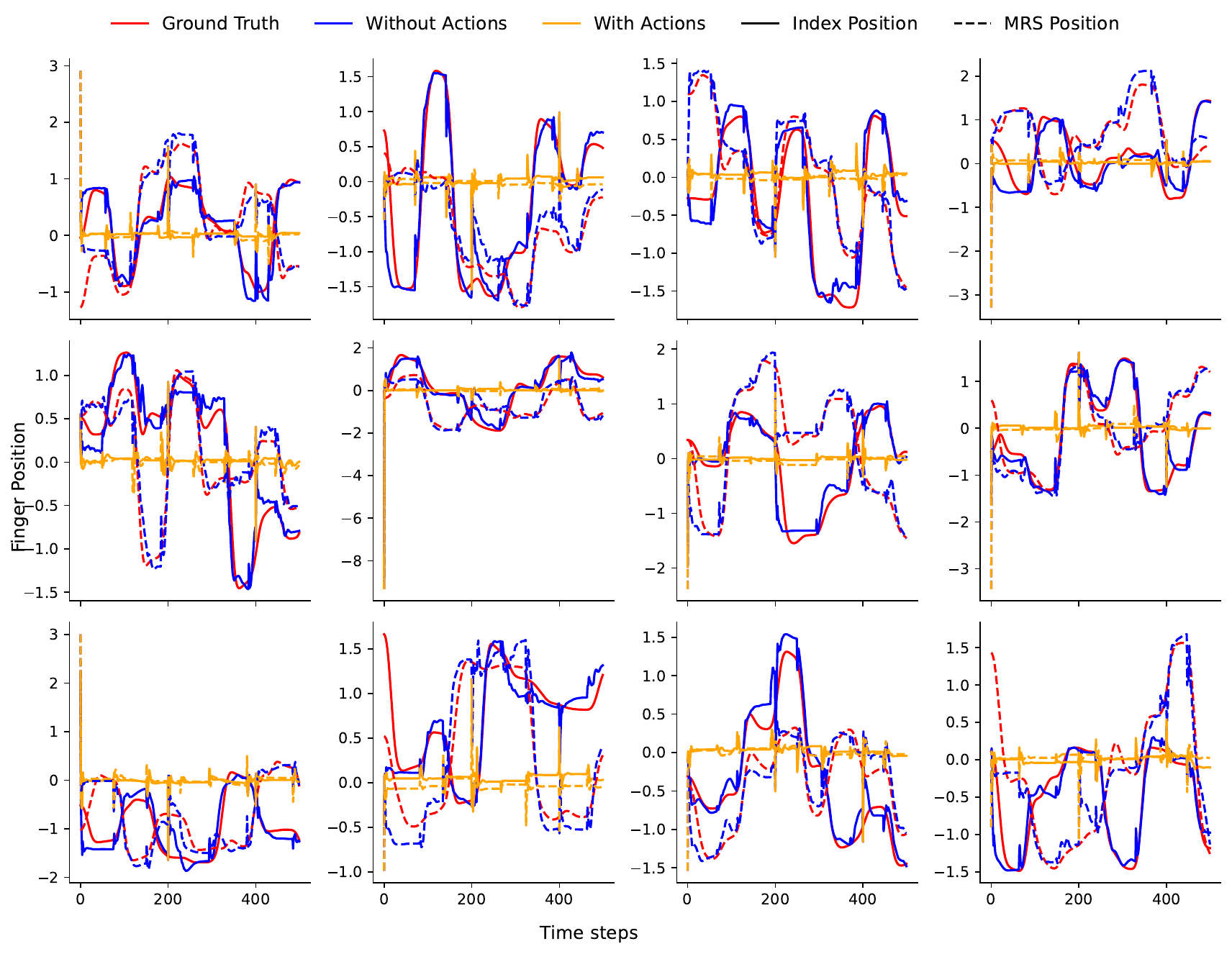}
    \caption{\textbf{Trajectories of  all LINK data subjects.}}
    \label{fig:sac_mc_multitask_example_trajectories_combined_width_112_height_86}
\end{figure}

\FloatBarrier
\section{DSR Training Protocol}
\label{sec:appendix_training}

The \gls{dsr} model is trained by minimizing the mean squared error between predicted and ground truth trajectories using the RAdam optimizer \citep{Liu2020On}, with subject-specific variance used for normalization. Training uses generalized teacher forcing, which linearly interpolates generated latent states with data-inferred states after loss computation. Since the latent state can be expanded with additional dimensions, the input dimension of the decoder does not necessarily correspond to the latent dimension and is therefore called the forcing dimension\citep{pmlr-v202-hess23a}. For non-identity observation models, losses are computed in both latent and observation space, and an additional penalty enforces invertibility of the observation model \citep{pmlr-v162-brenner22a}. Parameters are subject to $L_2$-regularization that pushes selected entries of $A$ toward 1 and entries of $W$ and $h$ toward zero \citep{schmidt2021identifying}. Subject vectors are initialized uniformly on $[-1, 1]$, and group-level parameters use
Xavier initialization \citep{pmlr-v9-glorot10a}. Reconstruction metrics are reported for the world model subsequently used for control. Further details on hierarchical \gls{dsr} training are given in \citet{ICLR2025_d4c96180}.

\subsection{Lorenz-63 data}
The Lorenz-63 system is a three-dimensional model of atmospheric convection that can exhibit stable fixed points, periodic orbits, and aperiodic chaotic dynamics depending on its parameters. Transitions between these regimes occur through bifurcations, including Hopf and homoclinic bifurcations. The system is defined by the coupled nonlinear ordinary differential equations
\begin{align}\label{eq:lorenz}
    \dot{x} &= \sigma (y-x)\\ \notag
    \dot{y} &= x (\rho -z)-y\\ \notag
    \dot{z} &= xy-\beta z
\end{align}
where 
$\sigma, \rho$ and $\beta$ are positive parameters. For \gls{dsr} training and testing, and for applying control to the true system, we integrate these equations using \texttt{odeint} from \texttt{scipy.integrate} at a step size of $0.01$
yielding $5000$ training and $100000$ test time steps per subject. Gaussian noise with a variance of 5\% of the subject-specific data variance is added to the training data. The 64 subjects correspond to the parameter configurations
$(\rho, \sigma, \beta)\in \{36, 66, 96, 126\}\times\{8, 9, 10, 11\}\times\{1, 2, 3, 4\}$ (\figref{fig:lorenz_3d_reconstruction_width_157_height_142}). The time series for all subjects were normalized together. For control, this transformation is reversed for the integration with the ODE.

\subsection{Lorenz-63 training and evaluation} \label{sec:appendix_lorenz}
Encoder and decoder are identity mappings. The three-dimensional observations are stacked with additional states obtained from a subject-specific linear mapping, which lifts them to the latent
dimension $L=26$. The ReLU nonlinearity acts on $P=12$ of these units, yielding at most $2^{12}=4096$ \glspl{lsr}.

\begin{figure}[!htb]
    \centering
    \includegraphics[width=\linewidth]{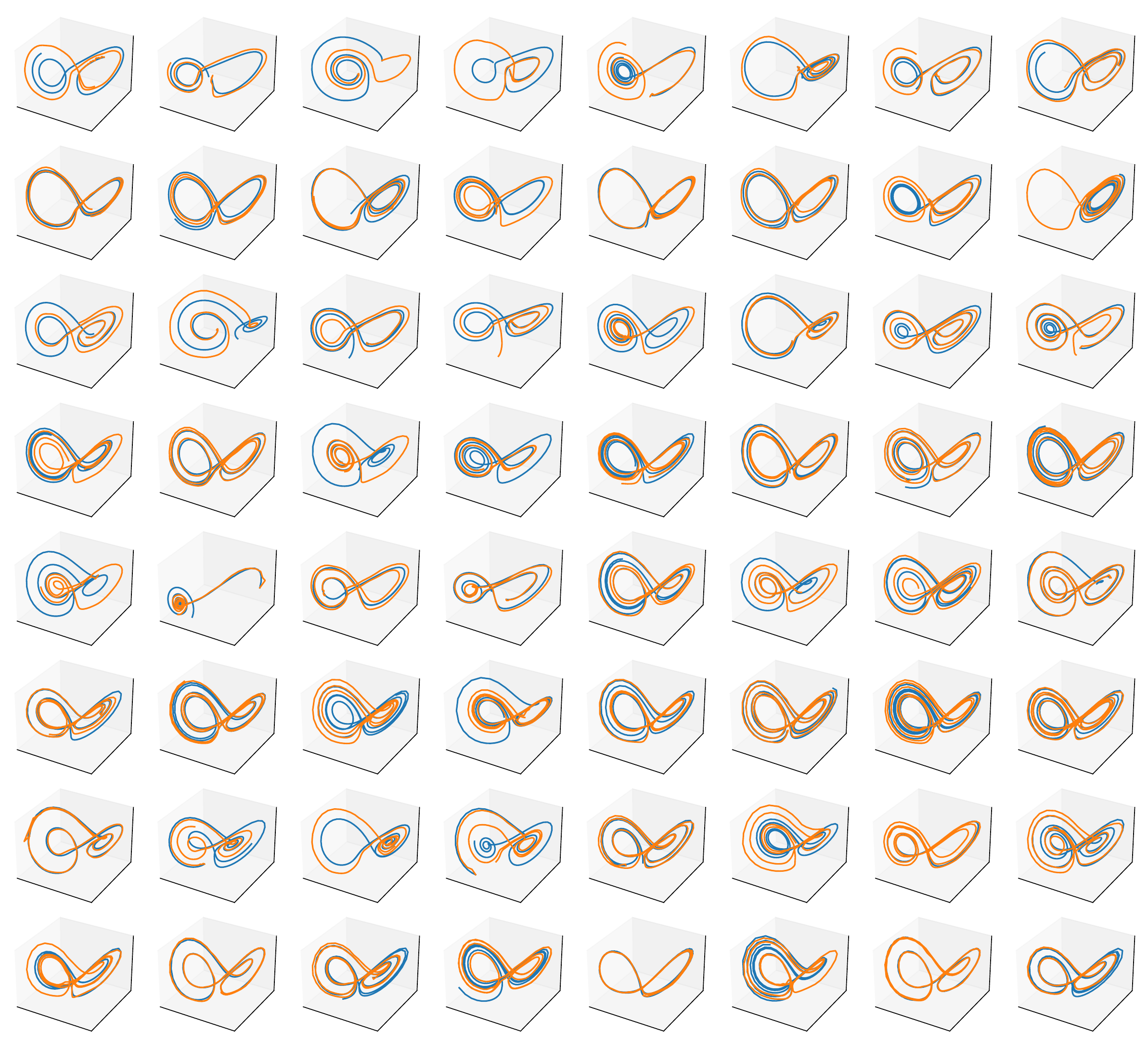}
    \caption{\textbf{Reconstruction of all Lorenz-63 subjects.} Ground-truth test trajectories (blue) and reconstructions (orange).}
    \label{fig:lorenz_3d_reconstruction_width_157_height_142}
\end{figure}

To assess the validity of the learned world model, we evaluate geometric agreement using the \gls{kl} divergence $D_{\mathrm{KL}}$ and temporal agreement using the dimension-averaged Hellinger distance $D_\mathrm{H}$ between power spectra \citep{10.1371/journal.pcbi.1007263,pmlr-v162-brenner22a,NEURIPS2022_495e55f3,pmlr-v202-hess23a}. With $p(x)$ the distribution of states in observation space, $f^i$ the power spectrum of dimension $i$, and $O$ the observation dimension, 
\begin{align}
    D_{\mathrm{KL}} &= \int p_\mathrm{gen}(x)\log\frac{p_\mathrm{gen}(x)}{p_\mathrm{true}(x)}dx\\
    D_\mathrm{H} &= \frac{1}{O}\sum\limits_{i=1}^O \sqrt{1-\int \sqrt{f_\mathrm{gen}^i(\omega)f_\mathrm{true}^i(\omega)}d\omega}.
\end{align}
 and can be estimated using discrete binning and Fast Fourier Transform, respectively. The  power spectra are smoothed using a Gaussian filter with standard deviation $\sigma = 20.0$ and cut off frequency $f_0=10\,$Hz for Lorenz-63 and $\sigma = 5.0$ and $f_0=10\,$ for the double-pendulum.\\
On the test set, the employed world model achieves median test-set values $D_{\mathrm{KL}}=1.8$ and $D_\mathrm{H}=0.15$ across subjects (\figref{fig:lorenz_3d_reconstruction_width_157_height_142}). For the 64 fine-tune subjects,  $D_\mathrm{H}=0.26$ remains comparable, while $D_{\mathrm{KL}}=5.3$ is higher.

\FloatBarrier
\subsection{Double-pendulum data}
Data for the double-pendulum are generated as for the Lorenz-63, using the equations of motion
\begin{align} \label{eq:pendulum}
    \dot{\theta}_1 &= \omega_1\\ \notag
    \dot{\theta}_2 &= \omega_2\\ \notag
    \dot{\omega}_1 &= \frac{1}{(m_1 + m_2) l_1 - m_2 l_1 \cos^2(\theta_2 - \theta_1)}(m_2 l_1 \omega_1^2 \sin(\theta_2 - \theta_1) \cos(\theta_2 - \theta_1) + \\ \notag 
     & m_2 g \sin(\theta_2) \cos(\theta_2 - \theta_1) + m_2 l_2 \omega_2^2 \sin(\theta_2 - \theta_1) - (m_1 + m_2) g \sin(\theta_1))\\ \notag
    \dot{\omega}_2 &= \frac{1}{(l_2 / l_1) ((m_1 + m_2) l_1 - m_2 l_1 \cos^2(\theta_2 - \theta_1))} (-m_2 l_2 \omega_2^2 \sin(\theta_2 - \theta_1) \cos(\theta_2 - \theta_1) + \\ \notag
     &(m_1 + m_2) g \sin(\theta_1) \cos(\theta_2 - \theta_1)- (m_1 + m_2) l_1 \omega_1^2 \sin(\theta_2 - \theta_1) - (m_1 + m_2) g \sin(\theta_2))
\end{align}
where $l_1$ and $l_2$ are the lengths of the two pendulum arms, $m_1$ and $m_2$ the masses at their ends, and $g=9.81$ the gravitational acceleration. The observations are then obtained as $x_t = (\sin(\theta_{1, t}), \cos(\theta_{1, t}), \sin(\theta_{2, t}), \cos(\theta_{2, t}), \omega_{1, t}, \omega_{2, t})$.

The kinetic and potential energies are
\begin{align}
    T &= \frac{1}{2} m_1 (l_1 \omega_1)^2 + \frac{1}{2} m_2 ((l_1 \omega_1)^2 + (l_2 \omega_2)^2 + 2 l_1 l_2 \omega_1 \omega_2 \cos(\theta_1 - \theta_2))\\ \notag
    V &= -(m_1 + m_2) g l_1 \cos(\theta_1) - m_2 g l_2 \cos(\theta_2).
\end{align}
We consider the 16 configurations $(l_1, l_2, m_1, m_2) \in \{0.5, 1\} \times \{1, 1.5\} \times \{0.5, 1\} \times \{1, 1.5\}$ for training and  $(l_1, l_2, m_1, m_2) \in \{0.1, 0.6\} \times \{0.5, 1.0\} \times \{0.25, 0.75\} \times \{0.75, 1.25\}$ for fine tuning. The training data comprise 50 time series of varying energy and 200 time steps per 
configuration, with initial angles and velocities sampled uniformly from [$0, 2\pi$] and [$-5, 5$], respectively. The test set consists of 10000 time steps per subject and training energy.

\FloatBarrier
\subsection{Double-pendulum training and evaluation}
\label{sec:appendix_pendulum_training}
Encoder and decoder are again identity mappings, with latent dimension $L=106$ and the nonlinearity applied to $P=84$ units. All other settings follow the Lorenz-63 experiments or are listed in Table \ref{hyperparameters}. For \gls{sac}, the reward weights each observation dimension by $\lambda_{reward} = (1, 10, 1, 10, 1, 1)$ to account for different scales and for the ambiguity of the sine, which is zero both at the target and at the opposite position. 
No further reward shaping, curriculum learning, or inductive bias for the Hamiltonian structure of the system is used. Actions act exclusively on the angular velocities $\omega_1$ and $\omega_2$.

A controlled trajectory counts as stabilized if, over the last 100 time steps, the mean absolute distance to the target is below $0.5$ rad for the angles and $5$ rad/s, for the angular velocities, and the standard deviation of the angular velocities is below $5$ rad/s (cf. \cite{gillen2020combiningdeepreinforcementlearning}).

\begin{figure}
    \centering
    \captionsetup[subfigure]{position=top, justification=raggedright, singlelinecheck=false}
    \begin{subfigure}{0.79\textwidth}
        \centering
        \caption{}
        \includegraphics[width=\linewidth]{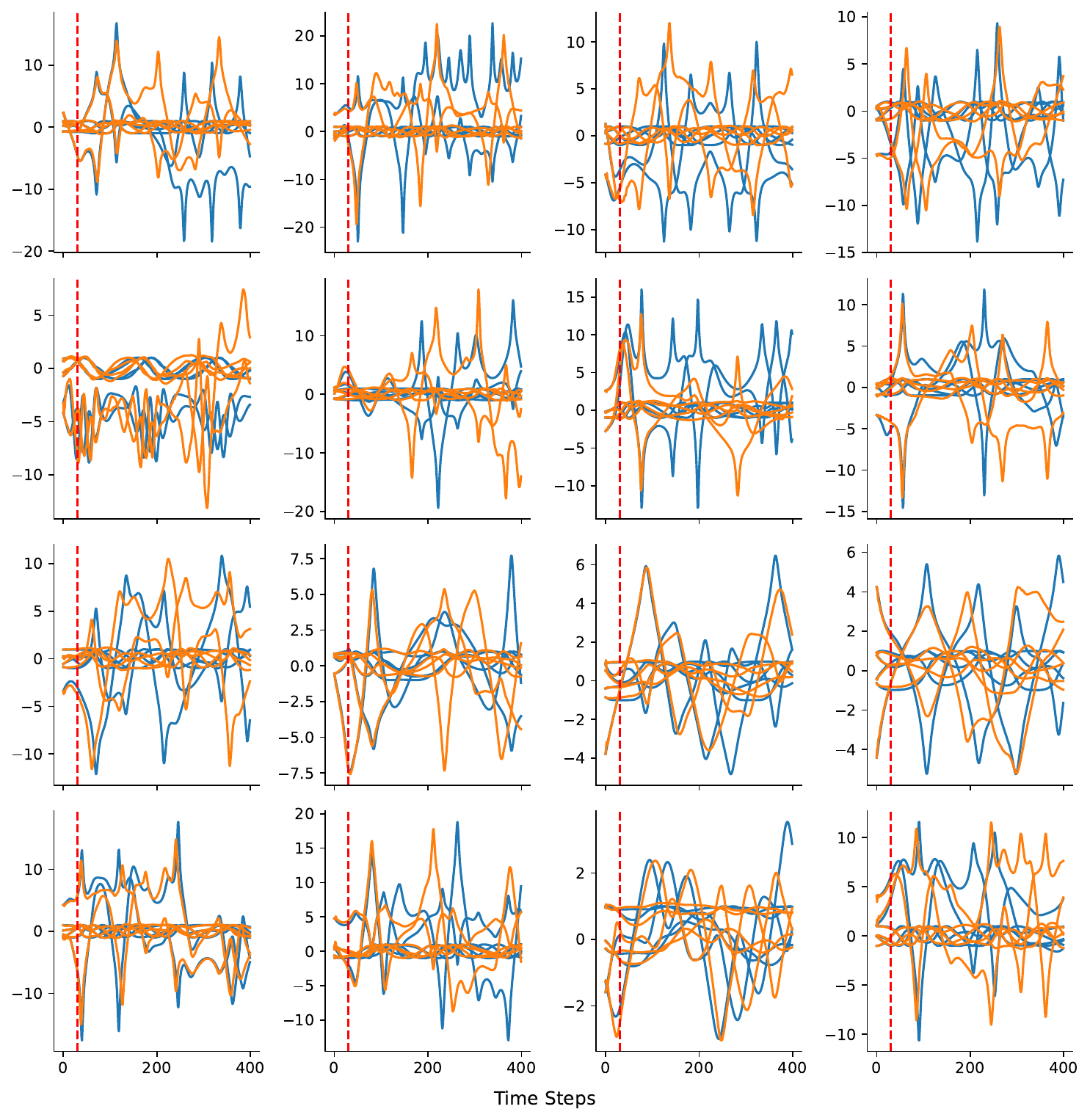}
        \label{fig:pendulum_reconstruction_all_width_96_height_99}
    \end{subfigure}
    \begin{subfigure}{0.19\textwidth}
        \centering
        \caption{}
        \vspace{9pt}
        \includegraphics[width=\linewidth]{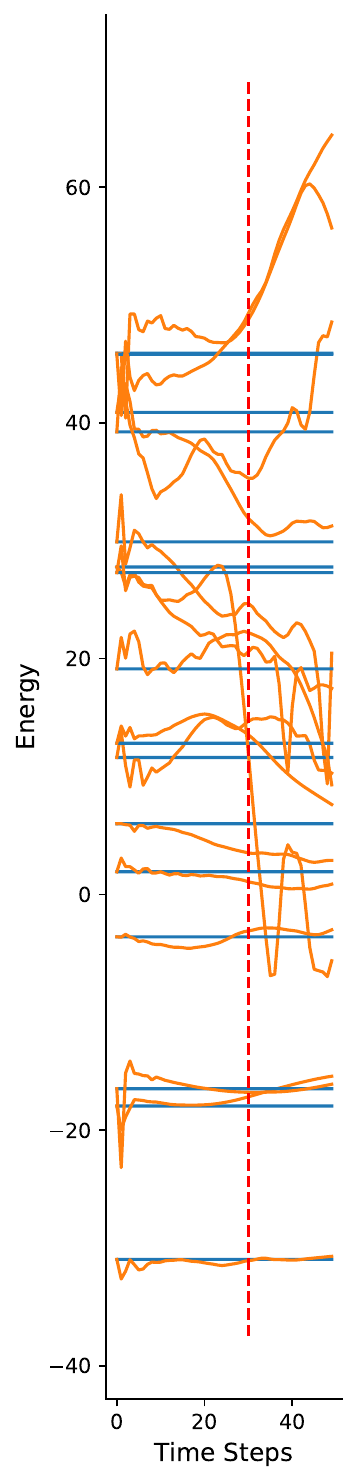}
        \label{fig:pendulum_energy_reconstruction_all_width_21_height_97}
    \end{subfigure}
    \caption{\textbf{Reconstruction of 16 double-pendulum subjects.} \textbf{a:} Ground truth (blue) and reconstructed (orange) trajectories of the $\theta_1, \theta_2, \omega_1$, and $\omega_2$ 
    agree closely up to the training sequence length of 30 time steps (red dashed line). \textbf{b:} Energy remains largely constant at first and fluctuates increasingly over time.}
\end{figure}

Due to its chaotic nature and the heterogeneity of the dynamics across subjects and energy levels, learning a world model for the double pendulum is a challenging task, and the prediction quality is limited
\citep{Chen2020Symplectic, bukowski2026predictionphysicslearningevaluating}. In terms of \gls{dsr} metrics, the model achieves a median value of $D_{\mathrm{KL}}=6.4$ and $D_\mathrm{H}=0.22$ (\figref{fig:pendulum_reconstruction_all_width_96_height_99}). These values are higher than for the Lorenz-63 system, although  different observation dimensions affect the scale of the \gls{kl} divergence.
Energy is largely conserved at first and fluctuates increasingly over time (\figref{fig:pendulum_energy_reconstruction_all_width_21_height_97}), a common consequence of accumulating prediction errors in learned physical models \citep{bukowski2026predictionphysicslearningevaluating, NEURIPS2022_495e55f3}. However, we emphasize that our model learns purely from data, without any physics-informed inductive biases.

\FloatBarrier

\subsection{LINK data}
\label{sec:appendix_link_data}
The Long-term Intracortical Neural activity and Kinematics dataset \citep{dandi:DANDI:001201_0.251023.2336} contains recordings from a single nonhuman primate performing a dexterous,
two-degree-of-freedom finger movement task. It includes 312 sessions recorded on 303 days over a period of about 3.5 years (1242 days), with data organized by sessions containing around 375 trials each.
With a sampling frequency of $50$\,Hz and recordings from 96 motor cortex channels of a Utah array, the observation dimension is 98 when including the two-dimensional finger positions.

We only considered trials with randomly selected targets. The spiking band power was first low-pass filtered using a 5th-order Butterworth filter with a cutoff frequency of 8\,Hz. Both finger position and velocity signals were smoothed using a Gaussian filter with a standard deviation of 8 time steps. Following filtering, all signals --- both neural and kinematic --- were normalized. Finally, the continuous time-series data was segmented into non-overlapping time series of 200 time steps each, forming the final input for our models.

\subsection{LINK training}
\label{sec:appendix_link_training}
We randomly selected 12 of the 312 sessions and treat them as subjects for our hierarchical approach. For each subject, the world model is trained and validated on average on 100 time series per subject and tested on 25. Both the neural data (spiking band power) and the kinematic variables are modeled using a Gaussian decoder. A causal convolutional encoder which performs a sequence of one-dimensional convolutions and whose inputs are padded on the left so that future observations do not influence the current latent state, infers the mean and variance of the latent state \citep{pmlr-v162-brenner22a}. In principle, encoders, decoders, and $C$ can also be learned hierarchically, although in our case they are learned for each subject. The external input to the latent \gls{alrnn} is the target position for both finger groups. The target state for \gls{sac} is 0 for both the index and the \gls{mrs} finger group.

On the test set, the median mean squared error across subjects is $0.81$ for the neural data and $0.09$ for the kinematic data (cf. \ref{fig:sac_mc_multitask_example_trajectories_width_55_height_49}).

\FloatBarrier
\subsection{Macaque delayed center-out reach data} \label{sec:appendix_mc}
In the delayed center-out reach task of \citet{PeiYe2021NeuralLatents}, a macaque controls a cursor to reach target positions in various maze configurations, while spiking activity of 182 neurons in primary motor and dorsal premotor cortex are recorded at 1000\,Hz together with hand and eye position. We consider only mazes without barriers. Spikes are binned at 50\,ms and hand and eye positions are normalized to $[-1,1]$, giving an observation dimension of 186.

\FloatBarrier
\subsection{Macaque delayed center-out reach training}
\label{sec:appendix_mc_maze_training}
The model receives the go cue and the goal position as external inputs to the latent \gls{alrnn}: the go cue as a single pulse and the target position during its presentation. A causal convolutional encoder is used and Hand and eye positions are each modeled with a two-dimensional Gaussian decoder. In contrast to LINK data, spike counts are modeled with a zero-inflated negative binomial decoder, which combines a negative binomial distribution for counts with a Bernoulli distribution for excess zeros and thus accounts for sparse, overdispersed data. The hand-position decoder is the sole goal decoder.

\begin{figure}[!htb]
    \centering
    \includegraphics[width=0.7\linewidth]{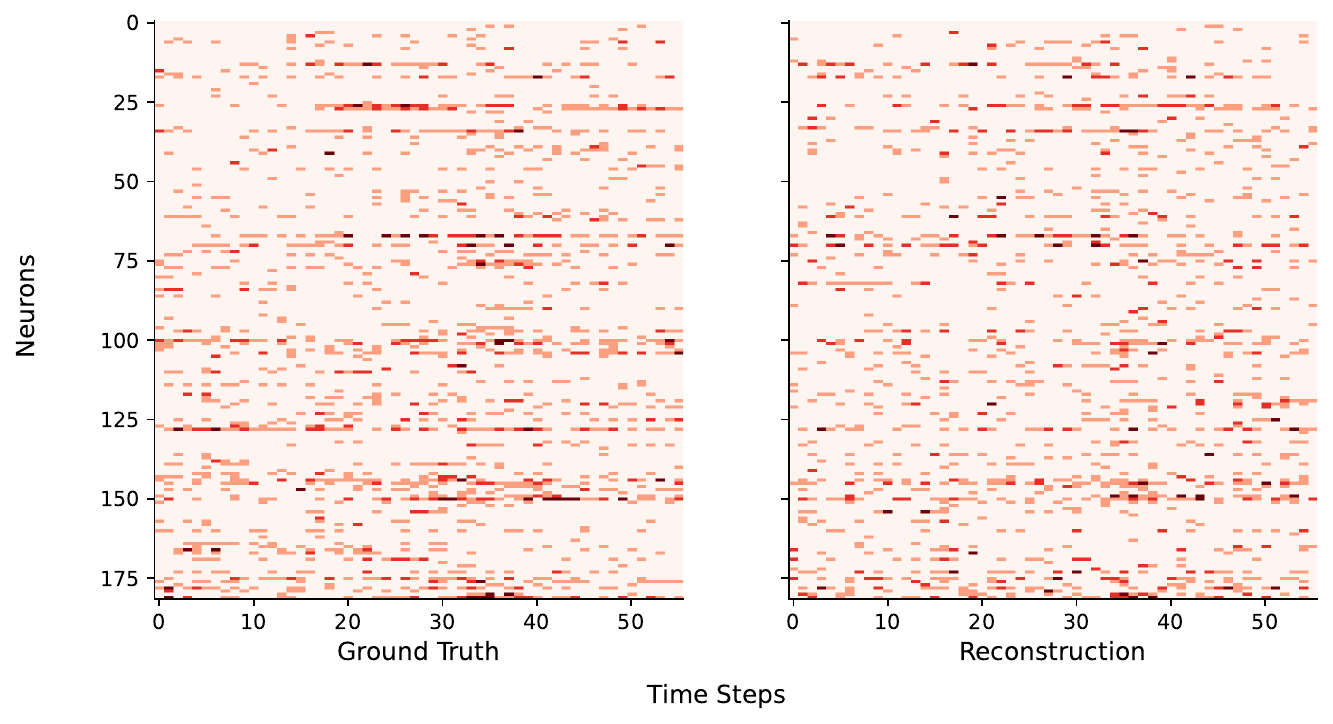}
    \caption{\textbf{Reconstruction of neural spike rates.} Empirical (left) and reconstructed (right) spike rates for a representative trial. Across all test trials, the world model achieves a variance-normalized mean squared error of $0.17\pm 0.06$.}
    \label{fig:mc_maze_all_spikes_width_86_height_46}
\end{figure}

Of 437 trials with an average duration of about 3\,s, 387 are used for training and validation and 50 are held out for testing. On the test set, the model achieves a mean squared error of $0.04 \pm 0.07$ for the position variables and a variance-normalized mean squared error of $0.17 \pm 0.06$ for spike rates (mean $\pm$ standard deviation across test trials; \figref{fig:mc_maze_all_spikes_width_86_height_46}).

\FloatBarrier
\section{SAC implementation}
\label{sec:appendix_sac}
The \gls{sac} implementation largely follows \citet{pmlr-v80-haarnoja18b}. The $Q$-functions and policy $\pi$ are ReLU feedforward networks whose weights and biases are hierarchized as described in the main text. Two independent $Q$-networks are trained, and the minimum of their predictions enters the gradient computation. Each $Q$-network has a target network updated by Polyak smoothing.
Since all rewards are non-positive, target values are clipped at a maximum of $0$, which prevents divergence of the value function. The policy is a Gaussian distribution, which permits the reparametrization trick.

The replay buffer is filled exclusively with trajectories generated by the world model (so-called \emph{imaginations}), and \gls{rl} gradients do not propagate back into the world model. To improve robustness on the true system, Gaussian noise is added to the latent state at each generated time step, with its magnitude treated as a hyperparameter (Table \ref{hyperparameters}). During evaluation, the expected values of both policy and the decoders are used. 
The temperature parameter is kept fixed and not optimized. The expectation over the decoder in the reward function (\eqref{eq:reward}) is approximated with a single sample.

To prevent actions from directly changing the outputs of goal decoders or other protected dimensions, a subject-specific projection maps latent actions onto the joint null space of the corresponding decoder weights. The relevant rows of the decoder mean weights are stacked, their null space is obtained via \gls{svd}, and an orthogonal projector onto this null space is formed. Actions are first $\tanh$-squashed to $[-1, 1]$ and then projected.

In experiments with external inputs, the policy additionally receives the current external input, which makes it time-dependent.

\section{Batching protocol}
We implement the training protocol as a measurement-aware batching hierarchy where each batch consists of multiple subbatches, with each subbatch containing samples exclusively from a single subject. 

For \gls{dsr}, all subbatches of a batch pass through the encoder and latent dynamics in a single forward pass. Losses are computed per subbatch, averaged, and combined with the global regularization terms. A single backward pass optimizer step then update all model parameters jointly. 

For \gls{sac}, training alternates between two phases. The actor phase collects exactly one batch per epoch, generates actions and rewards through environment interaction, and stores transitions in a replay buffer without gradient updates. The critic phase then samples multiple batches from this replay buffer and performs separate gradient updates
for $Q$-networks (minimizing 1-step Bellman errors) and policy (minimizing \gls{kl} divergence) using distinct optimizers, followed by Polyak averaging of target networks after each batch, effectively decoupling exploration from exploitation while maintaining the measurement-aware subbatch structure throughout both phases for hierarchical parameter learning.

\section{Baseline implementation}
\label{sec:appendix_baseline}
We compared the proposed hierarchical controller with three baselines that differed in how they use learned dynamics and access to controlled interactions. 

To ensure comparability, all methods were instantiated independently for each subject and used the same reward function, initial-state distribution, 500-step evaluation horizon, and projected action space. Raw actions were clipped and projected before execution. In the Lorenz-63 environment, only the projected $y$ and $z$ components were injected into the physical dynamics; in the double-pendulum environment, only the projected components corresponding to the two angular velocities were applied. Thus, all baselines respected the same constrained physical action interface as the proposed method.

CEM-\gls{mpc} \citep{pmlr-v155-pinneri21a} was implemented as a derivative-free receding-horizon planner that operates exclusively in the pretrained subject-specific \gls{alrnn} world model. At each evaluation step, candidate projected action sequences were simulated in the \gls{alrnn} and ranked by the same reward function as the proposed method, only the first action of the selected sequence was applied to the true system. Thus, CEM-\gls{mpc} required no additional controller training and did not use controlled true-system trajectories for planning. For CEM-\gls{mpc}, seeds were planner seeds controlling the stochastic sampling of candidate action sequences; no model or controller parameters were trained.

TD-\gls{mpc} followed the task-oriented latent-dynamics framework of \citet{pmlr-v162-hansen22a}. For each subject, TD-\gls{mpc} learned an encoder, latent transition model, reward predictor, twin $Q$-functions, and policy prior from controlled interactions collected in the true environment. Short-horizon action sequences were optimized in the learned latent model using an MPPI-style planner and evaluated with a learned terminal value estimate. The true environment was used only to execute the first selected action and provide subsequent observations; future true-system trajectories were never available to the planner.

The DreamerV3 RSSM agent \citep{Hafner2025} was implemented with categorical stochastic latents, observation, reward, and continuation prediction heads, and actor-critic learning through imagined trajectories. 
Like TD-\gls{mpc}, DreamerV3 learned exclusively from controlled true-environment interactions and was evaluated in closed loop, with each action selected from the learned model before being applied to the true system. The reported configurations were fixed before final evaluation. DreamerV3 included a restricted validation-search procedure, whereas TD-\gls{mpc} and CEM-\gls{mpc} used fixed configurations.

\section{Hyperparameters}
\label{sec:appendix_hypers}
In Table \ref{hyperparameters}, we report all hyperparameters which were either selected or optimized via Bayesian optimization. Learning rates and generalized teacher forcing coefficients follow exponential decay schedules.
For evaluation, we complete at least 6 \gls{sac} runs with different seeds, with group-level parameters initialized using Xavier initialization \citep{pmlr-v9-glorot10a}.

\begin{table}[!htb]
  \caption{Hyperparameters.}
  \label{hyperparameters}
  \begin{center}
    \begin{small}
      \begin{sc}
        \begin{tabular}{p{3.6cm}p{2.5cm}p{2.5cm}p{3cm}p{2.5cm}}
          \toprule
          & Lorenz-63 & Double\newline Pendulum & Macaque Delayed \newline Center-Out Reach & LINK\\
          \midrule
          \multicolumn{5}{c}{\textbf{DSR}} \\
          \midrule
          Latent Dim.              & $26$ & $106$ & $209$ & $94$\\
          Forcing Dim.            &3 & 6 & 10 & 10\\
          Subject Vector Dim.     & $4$ & $4$ & $1$ & $4$\\
          Sequence Length               & $30$ & $30$ & $60$ & $60$\\
          ReLU Units                    & $12$ & $84$ & $20$ & $29$\\
          Epochs                        & $5000$ & $4000$ & $3000$ & $3000$\\
          Learning Rate                 & $10^{-2}$ - $10^{-4}$& $3\cdot10^{-3}$- $8\cdot10^{-5}$ & $4.5\cdot10^{-4}$ - $4.1\cdot10^{-5}$& $5\cdot10^{-3}$ - $8\cdot10^{-9}$\\
          Teacher Forcing   & $6\cdot10^{-2} - 5\cdot10^{-2}$& $4\cdot10^{-2}$- $2\cdot10^{-3}$ & $0$ & $0$\\
          \midrule
          \multicolumn{5}{c}{\textbf{SAC}} \\
          \midrule
          Hidden Dim. $Q$     & $29, 25$ & $139, 117$ & $235, 135$ & $165, 149$\\
          Hidden Dim. $\pi$   & $20, 18$ & $146, 51$ & $198, 190$ & $83, 76$\\
          Action Penalty            & $0.1$ & $0.005$ & $0.0005$ & $0.01$\\
          Temperature               & $\exp(-0.5)$ & $\exp(3.9)$ & $\exp(0.1)$& $\exp(-4.53)$\\
          Environment Steps         & $14$ &  $38$& $59$& $41$\\
          Polyak Coefficient        & $8\cdot10^{-3}$ &  $1.6\cdot10^{-3}$ & $1.7\cdot10^{-3}$ &$1.7\cdot10^{-3}$\\
          Reward Discount           & $0.99$ & $0.99$ & $0.999$ & $0.999$\\
          Noise std    & $0.05$ & $0.21$ & $0$ & $0$\\
          Epochs                    & $5000$ & $5000$ & $5000$ & $5000$\\
          Learning Rate $Q$       & $6\cdot 10^{-4}$ - $2\cdot10^{-4}$ &$4.8\cdot 10^{-2}$ - $9\cdot10^{-6}$ & $4\cdot10^{-3}$ - $2\cdot10^{-4}$& $8\cdot10^{-3}$ - $3\cdot10^{-4}$\\
          Learning Rate $\pi$       & $2\cdot 10^{-4}$ - $5\cdot10^{-5}$ &$1.2\cdot 10^{-3}$ - $10^{-5}$ & $10^{-4}$ - $10^{-5}$& $10^{-3}$ - $2\cdot10^{-5}$\\
          \midrule
          \multicolumn{5}{c}{\textbf{Independent SAC}} \\
          \midrule
          Hidden Dim. $Q$     & $35, 28$ & $61, 54$ \\
          Hidden Dim. $\pi$   & $25, 25$ & $78, 72$ \\
          Action Penalty            & $0.1$ & $0.005$ \\
          Temperature               & $\exp(-0.5)$ & $\exp(-4.1)$\\
          Environment Steps         & $19$ &  $26$\\
          Polyak Coefficient        & $3\cdot10^{-3}$ &  $3.8\cdot10^{-3}$\\
          Reward Discount           & $0.99$ & $0.99$\\
          Noise std    & $0.03$ & $0.17$\\
          Epochs                    & $5000$ & $5000$\\
          Learning Rate $Q$       & $7\cdot 10^{-4}$ - $1.6\cdot10^{-4}$ &$8\cdot 10^{-3}$ - $5\cdot10^{-3}$\\
          Learning Rate $\pi$       & $1\cdot 10^{-4}$ - $2\cdot10^{-5}$ &$1.5\cdot 10^{-3}$ - $7\cdot10^{-4}$\\
          \bottomrule
        \end{tabular}
      \end{sc}
    \end{small}
  \end{center}
  \vskip -0.1in
\end{table}

\end{document}